\documentclass[journal]{IEEEtran}
\usepackage{cite}
\ifCLASSINFOpdf

\else
\fi

\usepackage{bbding}
\usepackage{times}
\usepackage{epsfig}
\usepackage{graphicx}
\usepackage{caption}
\graphicspath{{figures/}}
\usepackage{amsmath}
\usepackage{amssymb}
\usepackage[colorlinks,linkcolor=red]{hyperref}

\usepackage[utf8]{inputenc} 
\usepackage[T1]{fontenc}    
\usepackage{url}            
\usepackage{booktabs}       
\usepackage{amsfonts}       
\usepackage{nicefrac}       
\usepackage{microtype}      
\usepackage{array}
\usepackage{bm}
\usepackage{multirow}

\usepackage{times} 
\usepackage{helvet} 
\usepackage{courier} 
\usepackage{amsmath}
\usepackage{footnote}
\usepackage{multirow}
\usepackage{array}
\usepackage{algorithm}
\usepackage{algpseudocode}
\usepackage{amsfonts,amssymb}
\usepackage{booktabs}
\usepackage{color}
\usepackage{verbatim}

\begin{document}
	%
	\title{EGANS: Evolutionary Generative Adversarial Network Search for Zero-Shot Learning}
	%
	%
	%
	
	\author{
		Shiming~Chen,
		Shuhuang~Chen,
		Wenjin~Hou,
		Weiping~Ding,~\IEEEmembership{Senior~Member,~IEEE}
		and~Xinge~You,~\IEEEmembership{Senior~Member,~IEEE}
		
		\thanks{ This work was supported in part by the National Natural Science Foundation of China~(62172177, 61976120 ), the National Key R\&D Program of China (2022YFC3301004, 2022YFC3301003, 2022YFC3301704, 2022YFF0712300), the Natural Science Foundation of Jiangsu Province under Grant BK20191445, the Natural Science Key Foundation of Jiangsu Education Department under Grant 21KJA510004. Corresponding author: Weiping Ding (dwp9988@163.com) and Xinge You (youxg@hust.edu.cn) .}
		\thanks{Shiming Chen, Shuhuang Chen, Wenjin Hou and Xinge You are with the School of Electronic Information and Communication, Huazhong University of Science and Technology, Wuhan 430074, China.}
		\thanks{Weiping Ding is with the School of Information Science and Technology, Nantong University, Nantong, China.}
	}

	\markboth{IEEE Transactions on Evolutionary Computation}%
	{Shell \MakeLowercase{\textit{et al.}}: Bare Demo of IEEEtran.cls for IEEE Journals}
	
	\maketitle
	
	\begin{abstract}
		Zero-shot learning (ZSL) aims to recognize the novel classes which cannot be collected for training a prediction model. Accordingly, generative models (e.g., generative adversarial network (GAN)) are typically used to synthesize the visual samples conditioned by the class semantic vectors and achieve remarkable progress for ZSL. However, existing GAN-based generative ZSL methods are based on hand-crafted models, which cannot adapt to various datasets/scenarios and fails to model instability.  To alleviate these challenges, we propose evolutionary generative adversarial network search (termed EGANS) to automatically design the generative network with good adaptation and stability, enabling reliable visual feature sample synthesis for advancing ZSL. Specifically, we adopt cooperative dual evolution to conduct a neural architecture search for both generator and discriminator under a unified evolutionary adversarial framework. EGANS is learned by two stages: evolution generator architecture search and evolution discriminator architecture search. During the evolution generator architecture search, we adopt a many-to-one adversarial training strategy to evolutionarily search for the optimal generator. Then the optimal generator is further applied to search for the optimal discriminator in the evolution discriminator architecture search with a similar evolution search algorithm. Once the optimal generator and discriminator are searched, we entail them into various generative ZSL baselines for ZSL classification. Extensive experiments show that EGANS consistently improve existing generative ZSL methods on the standard CUB, SUN, AWA2 and FLO datasets. The significant performance gains indicate that the evolutionary neural architecture search explores a virgin field in ZSL.
	\end{abstract}


	%
	\IEEEpeerreviewmaketitle

	\section{Introduction and Motivation}\label{sec1}
	%
	%
	%
	%
	\begin{figure}[ht]
		\centering
		\includegraphics[scale=0.33]{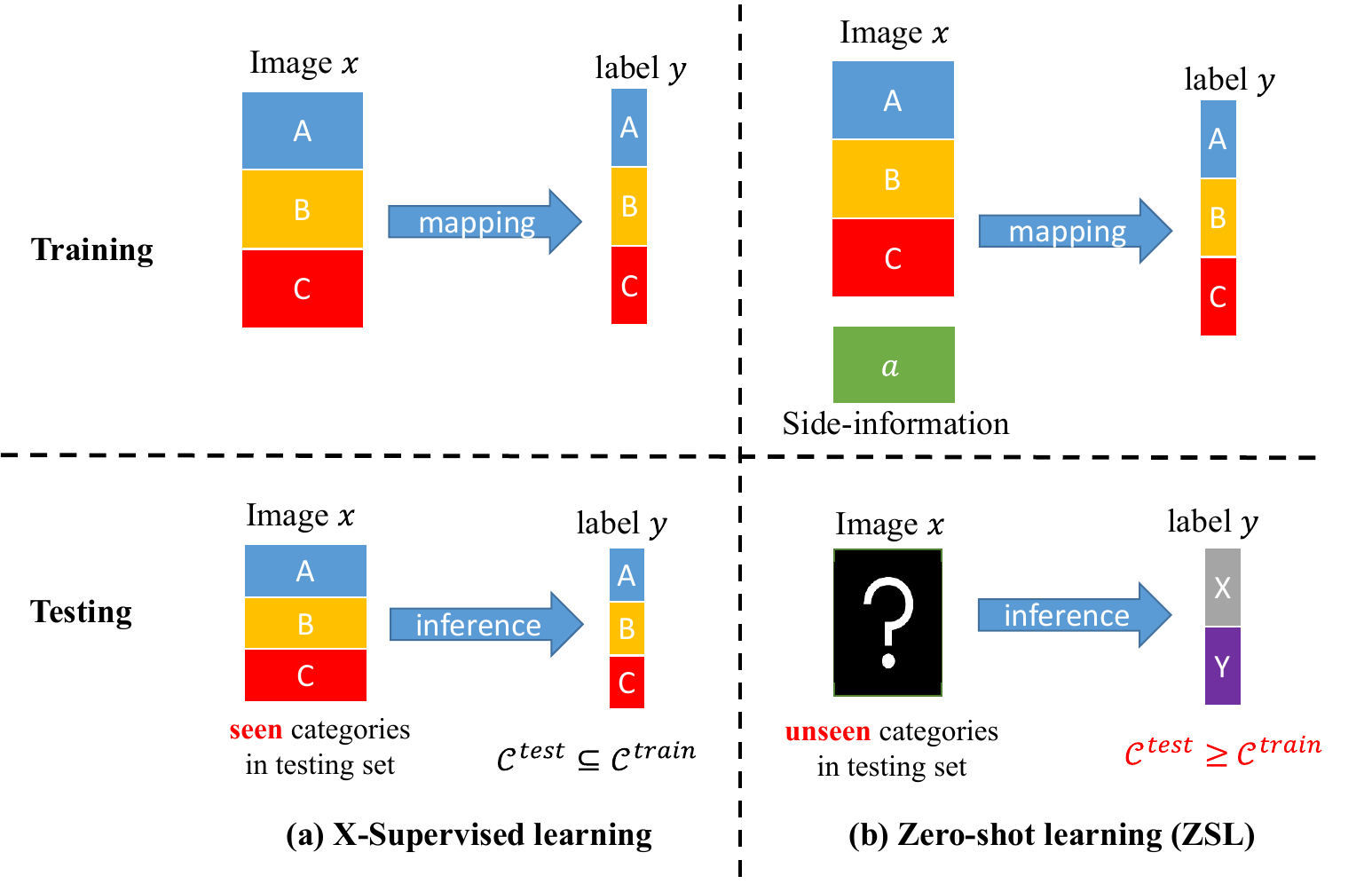}\vspace{-3mm}
		\caption{X-supervised learning \textit{VS} Zero-shot learning. (a) X-supervised learning (\textit{i.e.}, supervised/semi-supervised/unsupervised learning) recognizes the image of seen classes.  (b) Zero-shot learning recognizes the image of unseen classes by transferring semantic knowledge from seen classes to unseen ones.}
		\label{fig:intro-zsl}
	\end{figure}
	\IEEEPARstart{H}{uman} can learn novel concepts or objects based on prior knowledge without seeing them in advance. For example, given the clues that zebras appear like horses yet with black-and-white stripes of tigers, one can quickly recognize a zebra if he/she has seen horses and tigers before. Unlike humans, however, existing x-supervised machine learning methods (\textit{i.e.}, supervised/semi-supervised/unsupervised learning) can only classify samples belonging to the classes that have already appeared during the training phase (\textit{i.e.}, $\mathcal{C}^{test}\subseteq \mathcal{C}^{train}$, where $\mathcal{C}^{test}$ is the testing class set and $\mathcal{C}^{train}$ is the training class set), and they are not able to identify novel classes without any observed samples, as shown in Fig. \ref{fig:intro-zsl}(a). Motivated by this challenge, zero-shot learning (ZSL) was introduced to recognize new classes by transferring knowledge from seen classes to unseen ones based on the side-information\footnote{At present, most existing ZSL methods typically take the attribute descriptions as side-information, as attribute descriptions can provide informative semantic to capture the visual appearance.} (\textit{e.g.}, attribute descriptions, sentence embeddings)  \cite{Larochelle2008ZerodataLO,Palatucci2009ZeroshotLW,Lampert2009LearningTD,Lampert2014AttributeBasedCF,Fu2018ZeroShotLO,FuYanwei2015TransductiveMZ}, \textit{i.e.}, $\mathcal{C}^{test}\ge \mathcal{C}^{train}$, as shown in Fig. \ref{fig:intro-zsl}(b). Since ZSL is a foundational method of artificial intelligence, it is commonly applied in various tasks with wide real-world applications, \textit{e.g.}, image classification \cite{Frome2013DeViSEAD,Xian2019FVAEGAND2AF}, image retrieval \cite{Shen2018ZeroShotSH, Dutta2020SemanticallyTP}, semantic segmentation \cite{Bucher2019ZeroShotSS} and object detection \cite{Bansal2018ZeroShotOD}. Notably, ZSL is typically denoted as zero-shot image classification or visual object recognition \cite{Lampert2009LearningTD,Lampert2014AttributeBasedCF}, we also follow this standard in this paper. According to the various classification ranges during testing, ZSL methods can be grouped into conventional ZSL (CZSL) and generalized ZSL (GZSL). CSZL only predicts the objects of unseen classes, while GZSL predicts both seen and unseen classes  \cite{Xian2019ZeroShotLC}. Moreover, ZSL can also be categorized into inductive ZSL \cite{Xian2018FeatureGN,Xie2021GeneralizedZL}, which only utilizes the labeled seen data for training, and transductive ZSL \cite{Song2018TransductiveUE,Xie2021vman}, which takes both the labeled seen data and unlabeled unseen data for learning \cite{Xian2019ZeroShotLC}. Inductive ZSL is more reasonable and challenging in real life, we are thus focused on the inductive ZSL setting in this paper.

	Based on the side-information that is shared between the seen and unseen classes employed to support the knowledge transfer from seen classes to unseen ones, ZSL learns effective visual-semantic interactions between visual features and semantic features (extracted from side-information) with embedding-based methods, common space learning methods or generative methods. Embedding-based ZSL methods \cite{Akata2016LabelEmbeddingFI,Song2018TransductiveUE,Chen2022GNDANGN,Chen2022MSDN,Chen2021TransZero, Chen2021TransZeroCA} learn the embedding between seen classes and their class semantic vectors (\textit{i.e.}, visual$\rightarrow$semantic mapping), and then classify unseen classes using nearest neighbor search in the embedding space. Common space learning methods map the visual and semantic into a common space and recognize the unseen samples via embedding matching \cite{Frome2013DeViSEAD,Tsai2017LearningRV,Wang2017ZeroShotVR,Liu2018GeneralizedZL,Chen2021HSVA}. However, embedding-based and common space learning methods inevitably overfit to seen classes and limit their generalization from seen to unseen classes, since the embedding classification is simply learned on seen classes. To tackle this challenge, many generative ZSL methods have been introduced to synthesize visual feature samples for unseen classes using generative models (\textit{e.g.}, variational autoencoders (VAEs) \cite{Arora2018GeneralizedZL,Chen2021HSVA}, generative adversarial nets (GANs) \cite{Xian2018FeatureGN,Xian2019FVAEGAND2AF,Chen2021FREE}, and generative flows \cite{Shen2020InvertibleZR}) for data augmentation. As such, the generative ZSL methods compensate for the lack of training samples of unseen classes and convert ZSL into a supervised classification task. Accordingly, generative methods have achieved remarkable progress for ZSL and become a mainstream ZSL method.

	Thanks to the advantages of GANs synthesizing high-quality visual samples, a trend has emerged of synthesizing convolutional neural network (CNN) features of images using GAN architectures in ZSL \cite{Xian2018FeatureGN, Xian2019FVAEGAND2AF, Chen2023EvolvingSP, Zhao2022BoostingGZ, Hong2022SemanticCE}. Xian et al. \cite{Xian2018FeatureGN} initially introduced f-CLSWGAN to bring GANs techniques to the ZSL realm. To improve the training stability, Xian \cite{Xian2019FVAEGAND2AF} further proposed f-VAEGAN to combine the WGAN and VAE into a unified model. To progress this field, some improved methods enhance the semantic relevance for the synthesized visual samples of unseen classes using additional module and loss constraints, \textit{e.g.}, TF-VAEGAN \cite{Narayan2020LatentEF}, LisGAN \cite{Li2019LeveragingTI} and FREE\cite{Chen2021FREE}. Although the empirical success of these approaches, they all rely heavily on hand-crafted GAN architectures designed by human experts, resulting in poor adaptation on various datasets/scenarios. For example, one fixed GAN architecture cannot well adapt to the various datasets of animals (AWA2 \cite{Xian2019ZeroShotLC}), flowers (FLO \cite{Nilsback2008AutomatedFC}), and scene (SUN \cite{Patterson2012SUNAD}). Furthermore, the fixed architecture easily fails in the instability issue in GAN training. As such, automatically designing the GAN architectures with stable optimization customized for each specific ZSL task has become very necessary. 

	Recently, neural architecture search (NAS) has been proven to be effective for automatically searching superior network architecture with good generalization for various scenarios in various tasks \cite{Elsken2018NeuralAS, Ren2020ACS}, including the discriminative \cite{Liu2018DARTSDA,Chen2019DetNASNA} and generative tasks \cite{Gong2019AutoGANNA, Ying2021EAGANET}. To reduce search difficulty, the early NAS-based GANs \cite{Gong2019AutoGANNA, Wang2019AGANTA} search only generator with a fixed discriminator, resulting in a sub-optimal GAN architecture. As such, some recent works have simultaneously searched both generator and discriminator \cite{Gao2019AdversarialNASAN, Tian2020AlphaGANFD, Liu2021DynamicallyGG, Yan2021ZeroNASDG}. Based on the gradient optimization, Gao et al. \cite{Gao2019AdversarialNASAN} proposed the AdversarialNAS to search generator and discriminator simultaneously with an adversarial loss function. Meanwhile, Tian et al. \cite{Tian2020AlphaGANFD} introduced a fully differentiable search framework for the searching generator and discriminator of GAN. Since the architectures of the generator and discriminator in AdversarialNAS are deeply coupled, the search complexity and the instability of GAN training are increased. DGGAN \cite{Liu2021DynamicallyGG} alleviates instability by seeking the optimal architecture-growing strategy for the generator and discriminator, but it is time-consuming (\textit{e.g.}, taking 580 GPU days to search on the CIFAR-10 dataset). These NAS-based GAN methods show that automatically discovering the architectures of GANs will well improve the generalization and adaptation to various scenarios. Accordingly, we argue that NAS-based GANs will enable the generative ZSL methods to produce high-quality and generalized visual features of unseen classes on various datasets.

	Considering the success of evolutionary computation (EC) for the optimization GANs \cite{Schmiedlechner2018TowardsDC,Wang2019EvolutionaryGA,Costa2019COEGANET,Toutouh2019SpatialEG,Liu2020CatGANCG,Chen2020CDEGANCD} and NAS \cite{Stanley2002EvolvingNN,Liu2020ASO, Maziarz2018EvolutionaryNeuralHA,Huang2022ParticleSO,Xie2021BenchENASAB,Sun2020ANT,Ye2021ADF}, evolutionary neural architecture search (ENAS) based GANs models are introduced to improve the optimization and generalization of GANs recently \cite{Ying2021EAGANET, Lin2022EvolutionaryAS}. Although these methods can synthesize realistic and sharp images, they are not yet able to generate images of sufficient quality to train classifiers \cite{Xian2018FeatureGN,Xian2019FVAEGAND2AF}. Thanks to a large number of data in the supervised learning tasks, they are designed to search the optimal convolutional neural network (CNN) based on the pixel-level data, and cannot synthesize the reliable features samples based on the feature-level representations conditioned on object categories. As such, it is far from optimal to simply extend these methods to ZSL tasks for data augmentation. Furthermore, existing NAS-based GANs are limited by the specified adversarial optimization strategy, resulting in instability and sub-optimal GAN performance. 
		Motivated by the co-evolutionary algorithms that address GAN optimization pathologies effectively by evolving two populations: a population of generators and a population of discriminators  \cite{Schmiedlechner2018TowardsDC,Flores2022CoevolutionaryGA,Chen2020CDEGANCD,Costa2019CoevolutionOG,Costa2019COEGANET,Toutouh2020RepurposingHG}, we attempt to design an ENAS algorithm with cooperative dual evolution paradigm to automatically find superior multi-layer perceptron (MLP) architecture-based GAN models, which are effective in stable optimization and generalization for generative ZSL methods.

	In light of the above observation, we propose an evolutionary generative adversarial network search, dubbed \textit{EGANS}, to automatically design the generative network with good adaptation and stability, enabling reliable visual feature sample synthesis for advancing ZSL. Essentially, we adopt cooperative dual evolution to conduct an MLP architecture search (including the architecture and weight of the network) for both generator and discriminator under a unified evolutionary adversarial framework. EGANS is optimized with two stages: evolution generator architecture search and evolution discriminator architecture search. In the evolution generator architecture search, we fix the architecture of the discriminator and adopt a many-to-one adversarial training strategy to search optimal generator, \textit{i.e.}, the candidate generators and the fixed discriminator are in a many-to-one relationship. As such, there are multiple GAN models for optimization to search the optimal network architecture for generators, where the best generator is selected according to the fitness function. In the evolution discriminator architecture search, the best generator provides supervision signals for searching discriminators using a similar evolution search algorithm. Notably, we introduce the specific fitness function consisting of quality evaluation and architecture complexity to select the better offspring of the generator and discriminator for evolution. These two dual fitness functions encourage optimal cooperative dual evolution to optimize EGANS. Finally, we entail the searched GAN into a generative ZSL model to synthesize visual features of unseen classes for ZSL classification. Extensive experiments show that EGANS consistently improve existing generative ZSL methods on the standard datasets, \textit{e.g.}, CUB \cite{Welinder2010CaltechUCSDB2}, SUN \cite{Patterson2012SUNAD}, AWA2 \cite{Xian2019ZeroShotLC} and FLO \cite{Nilsback2008AutomatedFC}. These significant performance gains indicate that evolutionary neural architecture search effectively improves the generalization of generative ZSL methods and explores a virgin field in ZSL.

	To summarize, this study makes the main contributions as follows:
	
	\begin{itemize}
		\item  We propose a novel ENAS-based GAN, termed evolutionary generative adversarial network search (EGANS), to improve the adaptation and stability of generative ZSL methods. To achieve this goal, EGANS employs the cooperative evolution algorithm to conduct the evolution architecture search for the generator and discriminator.
		
		\item We design evolution generator architecture search and evolution discriminator architecture search, which evolve respectively by their own evolution search algorithms. Additionally, we also design two well-design fitness functions (\textit{i.e.}, simultaneously considering the quality evaluation and the architecture complexity) to select the better offspring during the evolution of the generator and discriminator.
		
		\item  We carry out extensive experiments on four benchmark datasets to demonstrate that our method consistently achieves significant improvement over the existing generative ZSL methods, which proves the superiority and great potential of EGANS.
	\end{itemize}

	The remainder of this paper is organized as follows. Section \ref{sec2} gives related works in the field of generative ZSL, neural architecture search (NAS), and evolutionary neural architecture search (ENAS). Our EGANS is illustrated in Section \ref{sec3}. The experiment evaluations are provided in Section \ref{sec4}. Section \ref{sec5} presents the discussion, and Section \ref{sec6} provides a conclusion and outlook.

	\section{Related Work}\label{sec2}
	
	\subsection{Generative ZSL}\label{sec2.1}
	Considering that the embedding-based ZSL method learns the embedding classifier only on seen classes, which leads the ZSL models to overfit to seen classes. To circumvent this challenge, generative ZSL methods apply the generative models (\textit{e.g.}, VAE and GAN) to synthesize the unseen visual features for data augmentation \cite{Arora2018GeneralizedZL,Xian2018FeatureGN,Shen2020InvertibleZR}. Accordingly, the generative ZSL methods have been proven to be effective in improving significant performance and become mainstream recently. Xian et al. \cite{Xian2018FeatureGN} introduced f-CLSWGAN to learn a generator to synthesize feature samples based on the WGAN \cite{Arjovsky2017WassersteinGA}. At the same time, Vinay Kumar Verma proposed SE-GZSL to synthesize image samples for unseen classes based on the VAE. Since i) the GAN is more powerful for synthesizing realm samples than VAE, and ii) Synthesizing the feature-level samples is easier than generating the high-dimension pix-level samples, f-CLSWGAN achieves better performance for ZSL. Thus, some subsequent generative ZSL methods are explored using feature-level GAN with MLP architecture. Specifically, Felix et al. \cite{Felix2018MultimodalCG} applied a semantic cycle consistency loss term into the f-CLSWGAN to enable good reconstruction of the original semantic features from the synthetic visual representations. f-VAEGAN \cite{Xian2019FVAEGAND2AF} is introduced to combine the advantage of VAE and GANs by assembling them into a conditional feature-generating model. LisGAN \cite{Li2019LeveragingTI} is employed to address the spurious and soulless generating problem of generative ZSL. Chen et al. \cite{Chen2021FREE} proposed a feature enhancement to alleviate the cross-dataset bias of generative ZSL based on the f-VAEGAN. Although these interesting methods have significantly improved the performance of generative ZSL, but neglect that the fixed GAN architectures are limited in the adaptation in various scenarios and stability of adversarial optimization. Orthogonal to these methods, we propose a novel ENAS-based GAN model to automatically search optimal GAN to synthesize reliable feature samples for generative ZSL. 
	
	\subsection{Neural Architecture Search}\label{sec2.2}
	According to the optimized used, existing NAS algorithms can be broadly grouped into three different categories: reinforcement learning (RL) based NAS methods \cite{Brock2017SMASHOM,Tan2018MnasNetPN}, gradient-based NAS methods \cite{Liu2018DARTSDA}, and evolutionary computation-based NAS (ENAS) methods \cite{Real2018RegularizedEF,Sun2020ANT}.  The RL-based methods depend on high-computation resources due to using tens of actions to get a positive reward, \textit{i.e.}, often requiring thousands of graphics processing cards (GPUs) with several days even on median-scale data set \cite{Ying2021EAGANET}. The gradient-based methods are more efficient than the RL-based methods. Since the ineffective relationship for adapting to gradient-based optimization, they typically find ill-conditioned architectures. Additionally, the gradient-based methods require constructing a supernet in advance, which also highly requires expertise and limits the capacity of NAS \cite{Bender2018UnderstandingAS,Dong2019OneShotNA}. ENAS simulates the evolution of species or the behaviors of the population in nature, to solve challenging optimization of NAS \cite{Liu2020ASO}.  Although NAS has achieved significant progress, they focus mainly on searching for optimal CNN architecture for the classification task. Furthermore, existing NAS methods train each discovered architecture individually to obtain their corresponding performance, resulting in substantial computational overhead. Ying \cite{Ying2021EAGANET} proposed a weight-sharing strategy to enhance search efficiency by constructing a large computational graph, in which each subgraph represents a neural network architecture. Consequently, all sub-network architectures can be evaluated without separate training by sharing weights within the large network. In our work, we adopt the weight-sharing method by constructing and training large generator and discriminator network architectures to search for the final architecture, thereby reducing the overhead associated with searching for network architectures.
	
	\begin{figure*}[ht]
		\centering
		\includegraphics[width=0.98\linewidth]{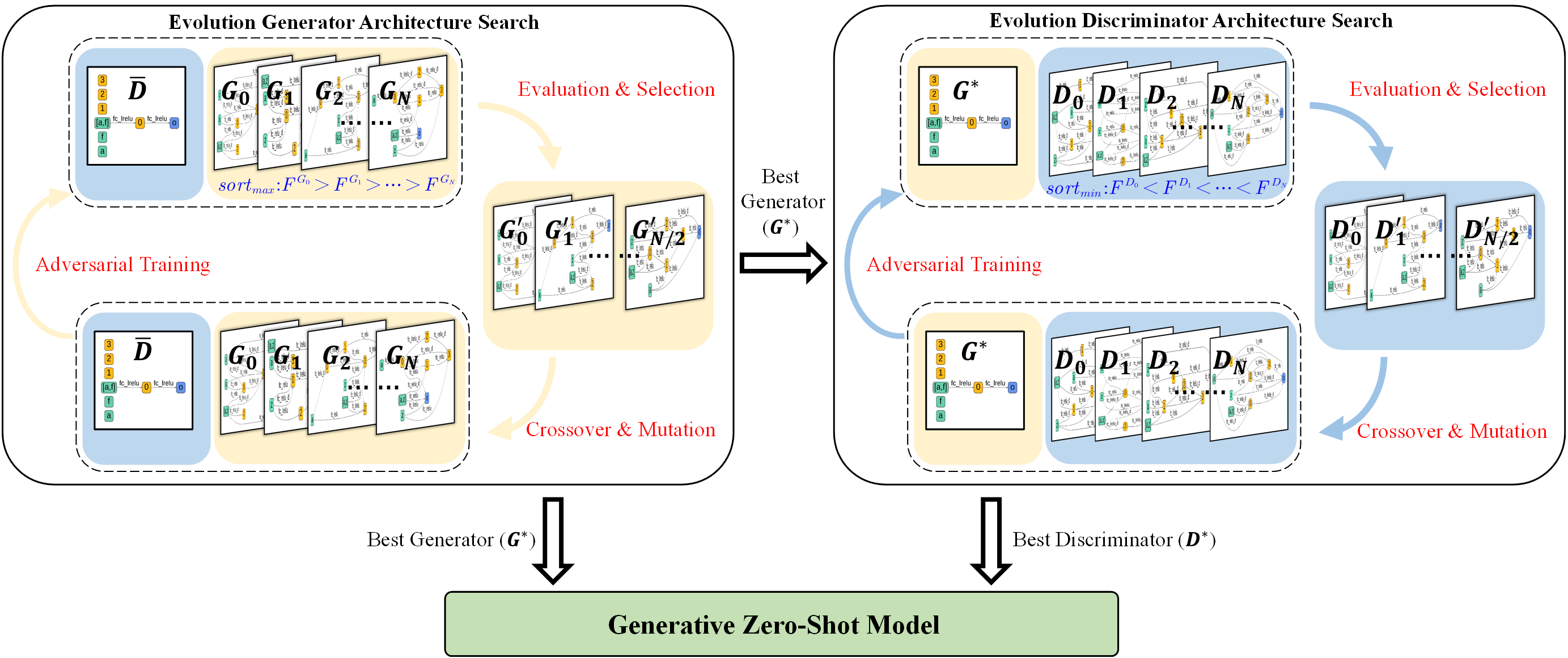}
		\caption{The pipeline of EGANS. In brief, EGANS consists of an evolution generator architecture search and an evolution discriminator architecture search, which is optimized by their evolutionary search algorithms. Once the optimal generator and discriminator are searched, they are entailed into the generative zero-shot model to synthesize visual feature samples of unseen classes to learn a supervised classifier, which is utilized for ZSL classification further.}
		\label{fig:pipeline}
	\end{figure*}
	
	\subsection{Evolutionary Neural Architecture Search}\label{sec2.3}
	The ENAS methods solve the NAS by exploiting effective evolution computation techniques \cite{Liu2021DynamicallyGG,Huang2022ParticleSO}, \textit{e.g.}, particle swarm optimization (PSO) and genetic algorithms (GAs).  PSO-based algorithms employ the particles to encode the network architectures and the velocity and position of each particle are updated toward the optimal network architecture in each iteration \cite{Sun2017APS}. Since GAs can tackle difficult problems that cannot be directly solved, many GA-based algorithms for the NAS have emerged recently \cite{Sun2017EvolvingDC,Assuno2018AutomaticEO,Lu2018NSGANetNA,Lu2020MultiObjectiveED}. Although some NAS methods have been introduced to automatically find network architectures for generative models (\textit{e.g.}, GANs) \cite{Gong2019AutoGANNA, Wang2019AGANTA,Tian2020AlphaGANFD, Liu2021DynamicallyGG}, ENAS-based GANs lack of enough attention \cite{Yan2021ZeroNASDG}. Ying \cite{Ying2021EAGANET} and Liu \cite{Lin2022EvolutionaryAS} initially introduce an evolutionary algorithm-based NAS framework to search GANs stably, but the models are based on the unconditional image generation task. Although the searched GAN architectures can unconditionally synthesize realistic and sharp images, these approaches are not capable
	for synthesize images of sufficient quality to train classifiers \cite{Xian2018FeatureGN}. As such, it is necessary to design effective ENAS-based GANs to synthesize feature-level samples for data augmentation in the ZSL task.
	
	\subsection{Co-Evolutionary Computation for GANs}\label{sec2.4}
		Co-evolutionary algorithms (coEA) overcome GAN optimization pathologies by evolving two populations: a population of generators and a population of discriminators  \cite{Schmiedlechner2018TowardsDC,Flores2022CoevolutionaryGA,Chen2020CDEGANCD,Costa2019CoevolutionOG,Costa2019COEGANET,Toutouh2020RepurposingHG}. Schmiedlechner \textit{et al.} \cite{Schmiedlechner2018TowardsDC} proposed Lipizzaner by training a two-dimensional grid of GANs with a distributed evolutionary algorithm. Subsequently, Lipizzaner is applied to medical image augmentations \cite{Flores2022CoevolutionaryGA}. In \cite{Toutouh2019SpatialEG}, Toutouh hybridized E-GAN \cite{Wang2019EvolutionaryGA} and Lipizzaner to combine mutation and population approaches to improve diversity GANs. Based on neuro-evolution and coevolution in the GAN training, Costa \textit{et al.} \cite{Costa2019COEGANET} devised COEGAN to provide a more stable training method and the automatic design of neural network architectures. To enable stable co-evolution between generator and discriminator, Chen et al. \cite{Chen2020CDEGANCD} developed CDE-GAN by conducting adversarial multi-objective optimization. However, the aforementioned CoEA-based GANs only take quality and diversity into consideration. In this work, we aim to design an ENAS algorithm to automatically find superior multi-layer perceptron (MLP) architecture-based GAN models with good generalization in the ZSL task. As such, we simultaneously take the quality and complexity into the fitness function to conduct stable co-evolutions with respect to the generator and discriminator.

	\section{Proposed Method}\label{sec3}

	\noindent \textbf{Problem and Notation Definition:} 
	First, we define some notations and the problem of ZSL. Given the seen class data $\mathcal{D}^{s}=\left\{\left(x_{i}^{s}, y_{i}^{s}\right)\right\}$ with $C^s$ seen classes, where $x_i^s \in \mathcal{X}^s$ is the $i$-th visual feature with 2048-dim extracted from a CNN backbone (\textit{e.g.}, ResNet101 \cite{He2016DeepRL}), and $y_i^s \in \mathcal{Y}^s$ is its corresponding class label. According to \cite{Xian2019ZeroShotLC}, $\mathcal{D}^{s}$ is divided into training set $\mathcal{D}_{tr}^{s}$ and test set $\mathcal{D}_{te}^{s}$. Another set $C^u$ is $\mathcal{D}_{te}^{u}=\left\{\left(x_{i}^{u}, y_{i}^{u}\right)\right\}$ consists of unseen classes, where $x_{i}^{u}\in \mathcal{X}$ are the visual features of unseen classes, and $y_{i}^{u} \in \mathcal{Y}^u$ is its corresponding labels. A set of class semantic vectors (semantic values annotated by humans according to attributes) of the class $c \in \mathcal{C}^{s} \cup \mathcal{C}^{u}$ with $|A|$ attributes, denoted as $a^c=\left[a_1^{c}, \ldots, a_{|A|}^{c}\right]^{\top} \in \mathbb{R}^{|A|}$. ZSL aims to learn a classifier, \textit{i.e.}, $f_{CZ S L}: \mathcal{X} \rightarrow \mathcal{Y}^{U}$ for CZSL, and $f_{G Z S L}: \mathcal{X} \rightarrow \mathcal{Y}^{U} \cup \mathcal{Y}^{S}$ for GZSL.

	\noindent \textbf{Framework Overview:} 
	Motivated by the success of the NAS algorithm in GAN \cite{Gong2019AutoGANNA, Wang2019AGANTA}, in this paper, we propose an \textit{evolutionary generative adversarial network search} (EGANS) circumvent drawbacks (\textit{i.e.}, adaptation and instability) of generative ZSL. In essence, EGANS incorporates dual evolution with respect to the generator and discriminator into a unified evolutionary adversarial framework to search for optimal MLP-like network architecture. As shown in Figure \ref{fig:pipeline}, EGANS takes an evolution generator architecture search and evolution discriminator architecture search to search the optimal generator and discriminator using their own evolution search algorithms. Then the best GAN model discovered by EGANS can be entailed into various generative ZSL models and adapted to various datasets/scenarios.
	
	\begin{figure*}[ht]
		\centering
		\includegraphics[width=0.98\linewidth]{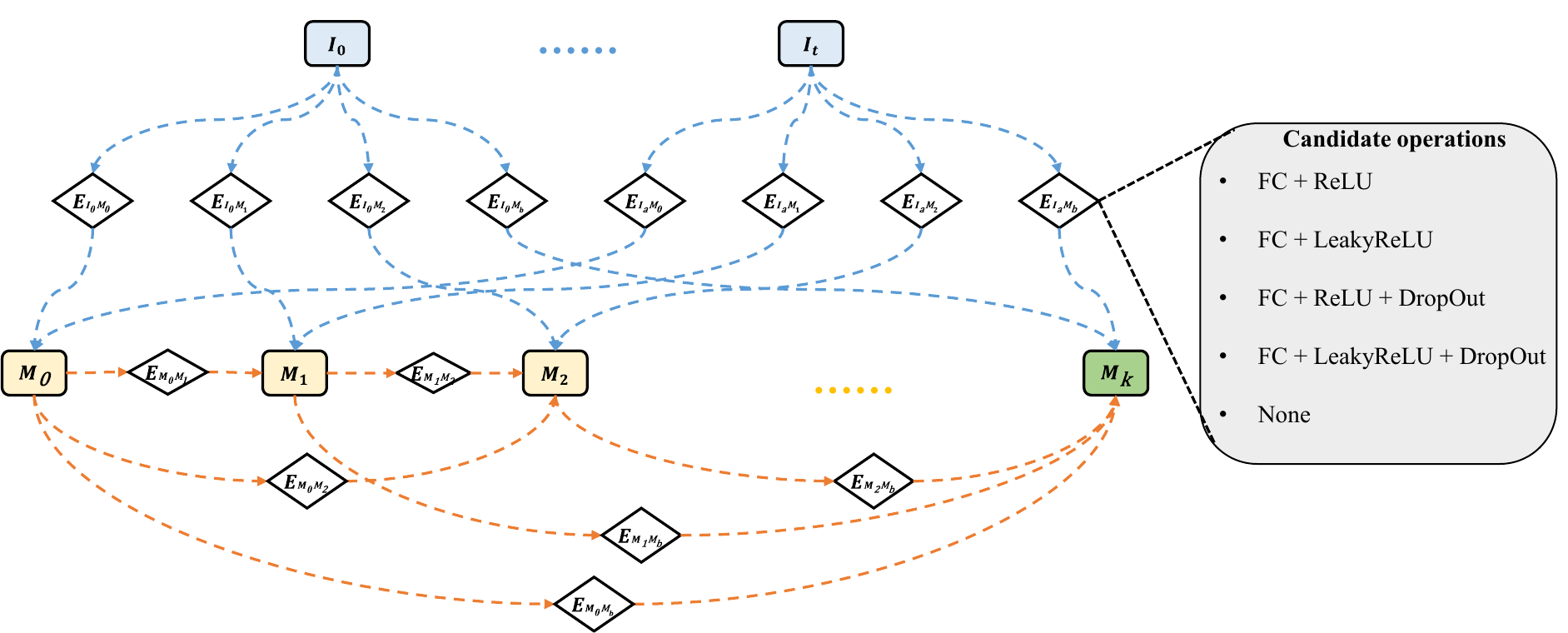}
		\caption{ The search space of our EGANS. The directed acyclic graph with input nodes $\{I_0,\cdots, I_j\}$, intermediate nodes $\{M_0,\cdots, M_{k-1}\}$ and
			output node $M_{k}$. (b) Each evolution operation is composed of predefined candidate operations used for connecting the nodes in a directed acyclic graph.}
		\label{fig:search-space}
	\end{figure*}

	\subsection{Preliminary Knowledge}
	
	\subsubsection{Revisiting GANs}
	
	A GAN \cite{Goodfellow2014GenerativeAN} adopts a unified framework to conduct the interaction between two different models, \textit{i.e.}, generator ($G$) and discriminator ($D$), which are used to solve a min-max game optimization problem. Given noisy vector $z \sim p_z$ as input, $G$ tries to learn real data distribution $p_{data}(x)$ (where $x$ can be image and features vector) by generating realistic looking samples $x \sim p_G(x)$ that is capable of fooling $D$ using $\min\limits_{G}$, while $D$ attempts to distinguish samples from the data distribution and the ones produced by $G$ using $\max\limits_{D}$. Formally, GAN \cite{Goodfellow2014GenerativeAN} can be formulated as:
	\begin{gather}
		\label{eq:gan}
		\min _{G} \max _{D} \mathbb{E}_{x \sim p_{\text {data }}}[\log D(x)]+\mathbb{E}_{z \sim p_{z}}[\log (1-D(G(z)))],
	\end{gather} 
	where $D(x)$ is the output of $D$ when the input is real data, and $D(G(z))$ is the output of $D$ when the input is the data produced by $G$.

	\subsubsection{Revisiting NAS}
	NAS aims to automatically search optimal network architecture and has achieved remarkable results in various fields \cite{Elsken2018NeuralAS, Ren2020ACS}. Mathematically, NAS can be formulated as a bilevel optimization problem:
	\begin{gather}
		\label{eq:nas}
		\begin{array}{ll} 
			& \beta^*=\arg \min \limits_{\beta} L_{\mathrm{val}}\left(\beta \mid w^*\right), \\
			\text { s.t. } & w^*=\arg \min \limits_{w} L_{\text {train }}(w \mid \beta),
		\end{array}
	\end{gather} 
	where $L_{\text {train }}$ and $L_{\mathrm{val}}$ are the training and validation loss, respectively; $w$ and $\beta$ are the weight and architecture of the neural network, respectively. This means that NAS tries to select the best architecture $\beta^*$ on the validation set, conditioned on the optimal network weights $w$ on the training set.

	\subsubsection{Revisiting Generative ZSL}
	In this work, we focus on generative ZSL \cite{Xian2018FeatureGN,Xian2019FVAEGAND2AF}. Generative ZSL typically learns a conditional generator G: $\mathcal{A} \times \mathcal{O}\rightarrow \tilde{\mathcal{X}}$, which takes the class semantic vectors/prototypes $a\in \mathcal{A}$ and Gaussian noise vectors $z\in \mathcal{Z}$ as inputs, and synthesizes the class-specific visual feature samples $\tilde{x}\in \tilde{\mathcal{X}}$. After training, the generator is utilized to synthesize a large number of class-specific visual feature samples for unseen classes according to the corresponding semantic vector $a^{u}$. Finally, we can take $\{\tilde{\mathcal{X}}^u , \mathcal{X}^s\}$ to learn a supervised classifier (\textit{e.g.}, softmax) over $y\in\mathcal{Y}$ for ZSL classification. As such, the ZSL task is converted to a supervised classification problem.

	\subsection{Search Space}\label{sec3.1}
	To explore more hierarchical network architectures, we establish a highly flexible search space for both the generator and discriminator. As shown in Fig. \ref{fig:search-space}, the generator and discriminator networks can be represented as directed acyclic graphs $G_g\left(\mathcal{N}_g, E_g\right)$ and $G_d\left(\mathcal{N}_d, E_d\right)$, respectively, where $\mathcal{N}=\{I_0,\cdots, I_t, M_0, \cdots, M_k\}$ is the set of nodes and $E$ is the edge between two nodes. Each node represents a feature vector with a specific dimension, and each edge $E_{{\mathcal{N}_i}{\mathcal{N}_j}}$ represents the transformation of ${\mathcal{N}_i}$ into ${\mathcal{N}_j}$ through a candidate operation. 
	
	$G_g\left(\mathcal{N}_g, E_g\right)$ includes three input nodes, namely the semantic vector $a$, the noise vector $z$, and the concatenated vector $a+z$. Additionally, the acyclic graph includes five ordered nodes to avoid over-complex models, \textit{i.e.}, $M_k$, where $(k=4)$ and node $M_4$ is the output node. Considering using ResNet as the backbone network, the dimensions of the five nodes can be set to 512, 1024, 2048, 4096, and 2048, respectively. Each edge in $G_g\left(\mathcal{N}_g, E_g\right)$ shares a common set of candidate operations. However, only one operation is allowed to be activated per edge. In the MLP networks, fully connected layer (FC), activation functions (e.g., ReLU and LeakyReLU), and dropout function are the basic units. To accommodate the ZSL task, our candidate operations include:
	
	\begin{itemize}
		\item FC+ReLU    
		\item FC+ReLU+DropOut
		\item FC+LeakyReLU    
		\item FC+LeakyReLU+DropOut 
		\item None    
	\end{itemize}
	
	Here, “FC” represents a fully connected layer, “ReLU” and “LeakyReLU” are two activation functions, “DropOut” is a regularization technique, and “None” indicates that there is no connection between the two nodes. We use one-hot encoding for each edge. For example, the [0, 1, 0, 0, 0] of edge $E_{I_1M_1}$ indicates that the \textit{FC+ReLU+DropOut} operation is activated. This directed acyclic graph contains 8 nodes and 25 edges, resulting in a search space containing $5^{25}$ sub-architectures. The "None" operation enables sub-architectures with different network dimensions and layers.
	
	$G_d\left(\mathcal{N}_d, E_d\right)$ includes three input nodes, which are the class semantic vector $a$, the visual feature vectors $x$, and the concatenated vector $a+x$. Similarly, taking a model with ResNet as the backbone network as an example, the dimensions of the five ordered nodes can be set to 4096, 2048, 1024, 512, and 1, respectively. The candidate operations are the same as those in the generator network.
	
	EGANS employs a weight-sharing training strategy, which involves a large computational graph containing both the generator and discriminator, namely Large\_Generator (\textit{i.e.}, $G_L$) and Large\_Discriminator (\textit{i.e.}, $D_L$). This is formed by activating all edge operations in the aforementioned directed acyclic graphs, \textit{i.e.}, $G_g\left(\mathcal{N}_g, E_g\right)$ and $G_d\left(\mathcal{N}_d, E_d\right)$.

	\subsection{Neural Architecture Search}\label{sec3.3}
	Following iterative optimization in adversarial training, we conduct cooperative dual evolution for stable optimization of our method, including \textit{evolution generator architecture search} and \textit{evolution discriminator architecture search}. Since CLSWGAN is the most popular baseline, all the generative ZSL methods adopt the same discriminator architectures as CLSWGAN. As such, we take the discriminator of CLSWGAN for the initialization of EGANS's discriminator. As illustrated in Fig. \ref{fig:pipeline}, we first fix the initialized discriminator $\bar{D}$ and evolve to search for the optimal generator architecture $G^*$ via evolution generator architecture search. Based on this optimal generator architecture $G^*$, we further search for the discriminator architecture $D^*$ that adapts to the generator well via evolution discriminator architecture search. Thus, the searches of the generator and discriminator are based on the cooperative evolution algorithm. Finally, we combine $G^*$ and $D^*$ to form the optimal GAN, which serves as the generative network for generative zero-shot learning.

	\subsubsection{Evolution Generator Architecture Search}
	
	As shown in Fig. \ref{fig:pipeline} (left), we search for the best generator $G^*$ with a fixed discriminator $\bar{D}$, whose architecture is the same as CLSWGAN \cite{Xian2018FeatureGN}. Specifically, the fixed discriminator $\bar{D}$ consists of architecture and weights variables, \textit{i.e.}, $\bar{D}\left(\bar{\beta},w_{\bar{D}}\right)$. Based on the evolutionary algorithms, we take \textit{Adversarial Training}, \textit{Evaluation \& Selection}, and \textit{Crossover \& Mutuation} for evolving generator architecture search. We first randomly generate $N$ individual generators $\{G_1, \dots, G_N\}$, and form the generator population $P_G$.  Each $G_i$ can be represented by a combination of architecture and weights variables $G_i\left(\alpha,w_{G_i}\right)$. $N$ individual generators are combined with the same discriminator $\bar{D}$ to form $N$ GAN models, which are optimize with a weight-sharing and many-to-one adversarial training strategy. The optimizing objective of the evolution generator architecture search can be formulated as:
	\begin{gather}
		\label{searchG}
		\alpha^{*}=\arg \max _{\alpha_{i}}\left\{F_{G}\left(\alpha_{i} \mid w_{G_{i}}^{*}, w_{\bar{D}}^{*}, \bar{\beta}\right), i \in\{1, \ldots, N\}\right\},
	\end{gather}
	\begin{gather}
		\label{trainG}
		\begin{array}{ll} 
			\begin{aligned}
				\text { s.t. } w_{G_{i}}^{*}=&\arg \min _{w_{G_{i}}} \mathbb{E}_{z \sim p(z)}\left[-\bar{D}\left(a, G_{i}(a, z)\right)\right] \\
				w_{\bar{D}}^{*}=&\arg \max _{w_{\bar{D}}} \sum_{i=1}^{N} \mathbb{E}_{x \sim p_{\text {data }}(x)}[\bar{D}(a, x)] \\
				&-\mathbb{E}_{z \sim p(z)}\left[\bar{D}\left(a, G_{i}(a, z)\right)\right],
			\end{aligned}
		\end{array}
	\end{gather}
	where Eq. \ref{trainG} corresponds to the adversarial training of the generator population and the discriminator.
	
	During the evolution, we take the selected $N/2$ individuals as parents with the fitness function $\mathcal{F}_G$ to perform crossover and mutation to produce additional $N/2$, forming a new population of $N$ offspring. The next round of training and evolution continues until the optimal model is discovered. We provide a detailed description of the evolution generator architecture search algorithm below.
	
	\noindent \textbf{Many-to-One Adversarial Training:} 
	In each round, we generate $N$ candidate generators to form the generator population $P_G$. All candidate generators share the weights $W_{G_L}$ of the Large\_Generator $G_L$. And the first round's candidate generators are randomly generated, we use a 50\% probability to change the candidate generators' edge operations to "None" (disconnecting the connection of two nodes), avoiding the initial architecture being overly complex.
	
	We then train these $N$ GANs using a many-to-one strategy, \textit{i.e.}, optimizing the parameters of the $N$ generators and the fixed-architecture discriminator with Eq. \ref{trainG}. These $N$ GAN models share the same discriminator and undergo training for multiple epochs per round. To make fair comparisons among generators, we uniformly select one generator from the $N$ candidate generators for each training batch along with the fixed discriminator. In essence, for each training batch, we activate and train the parameters $w_{G_i}$ corresponding to the $G_i$ architecture within $G_L$, along with the parameters $w_{\bar{D}}$ for the discriminator $\bar{D}$. The many-to-one training mechanism offers two benefits: i) the fixed discriminator $\bar{D}$ can be trained with various generators, which can be regarded as an ensemble method to some extent, avoiding $\bar{D}$ overfitting and becoming significantly stronger than the generators; ii)  we can fairly compare their performance to find the optimal generator, as different generators cooperate with the same discriminator for training.
	
	\noindent \textbf{Evaluation \& Selection:}
	After the adversarial training, we introduce a fitness function $F_G$ to evaluate the quality of individuals, and then select $N/2$ higher-quality individuals as parents for the next evolution. $F_G$ includes two fitness components, \textit{i.e.}, the quality fitness $F_{G^q}$ and the architecture complexity $F_{G^c}$. We employ $F_{G^q}$ to evaluate the generation quality of the generator, formulated as:
	\begin{gather}
		\label{eq:F_Gq}
		\mathcal{F}_{G^q}=\mathbb{E}_{z \sim p(z)} \bar{D}\left(a, G_{i}(a, z)\right).
	\end{gather}
	The higher quality obtained by the generators, the more realistic visual feature samples it produces. During training, the discriminator is continuously upgraded to reach an optimal state and provides the quality evaluation for generators at each evolutionary step. Additionally, considering the diversity of architecture within our search space, we aim to ensure that the selected architecture balances training efficiency. To achieve this, we introduce architecture complexity $\mathcal{F}_{G^c}$ to evaluate the generator architecture{\color{blue}:}
	\begin{gather}
		\label{eq:F_Gc}
		\mathcal{F}_{G^c}= C_{E_{nn}}(G_i)/C_{E_{all}}(G_i),
	\end{gather}
	where $C_{E_{nn}}(\cdot)$ is a function for counting edges without using the “None” operation, and $C_{E_{all}}(\cdot)$ is a function for counting the full edges in the whole graph. When an evolved $i$-th generator architecture is more complex, ${F}_{G_i^c}$ is larger. 
	
	Accordingly, $\mathcal{F}_{G}$ is formulated as:
	\begin{gather}
		\label{eq:F_G}
		\mathcal{F}_{G}=\mathcal{F}_{G^q} - \lambda_G\times \mathcal{F}_{G^c}.
	\end{gather}
	$\lambda_G > 0$ is a weight to balance generation quality and the architecture complexity of the generator. To this end, the performance $\mathcal{F}_{G_i}$ of each evolved offspring $G_i$ is evaluated using $\mathcal{F}_{G}$. Generally, the larger the fitness value $\mathcal{F}_{G}$, the better the generation performance with the higher training efficiency. Finally, a simple yet useful survivor selection strategy $(\mu, \lambda)$-selection \cite{Kramer2016Machine} is applied to select the new parents of the next evolution according to the fitness value of existing individuals. The selection function for the offspring of generator evolution is defined as:
	\begin{gather}
		\label{eq:select-G}
		\{G'_1,\cdots,G'_{N/2}\}\leftarrow sort_{max}\{\mathcal{F}_{G_i}\}_{i=1}^N.
	\end{gather}
	As such, we select the top $N/2$ individuals $\{G'_1,\cdots,G'_{N/2}\}$ to serve as the parents for the next generation. Interestingly, the $\mathcal{F}_{G}$ and $\mathcal{F}_{D}$ are duality, which encourage the stable optimization for EGANS.
	
	\noindent \textbf{Crossover and Mutation:}
	We set the $N/2$ individuals obtained from the selection step as parents and then perform crossover and mutation on these $N/2$ parents with a 50\% probability, respectively, until $N/2$ new individuals are generated. As described in the search space section, the architecture of each subnetwork is encoded by a set of one-hot sequences, where each one-hot sequence represents an edge, and the position of “1” indicates the activated candidate operation on that edge. 
	
	Consequently, the basic units of crossover and mutation are one-hot sequences. For crossover, we randomly select two parents and exchange half of the edge operations selected randomly, \textit{i.e.}, recombining half of the one-hot sequences and then removing redundant nodes. For mutation, we randomly replace the connection operations for 1/2/4 edges with the probabilities of 50\%/30\%/20\%, respectively. This means we randomly select 1/2/4 one-hot sequences and change their position of “1”.
	
	\subsubsection{Evolution Discriminator Architecture Search}
	
	As illustrated in Fig. \ref{fig:pipeline} (right), we fix the generator architecture once the optimal generator $G^*$ is discovered in the evolution generator architecture search. Then we adopt the same strategy as in the evolution generator architecture search to discover the optimal discriminator architecture. In the evolution discriminator architecture search, we first randomly generate $N$ discriminator individuals $\{D_1, \dots, D_N\}$ and form discriminator population $P_D$. Each $D_i$ is the combination of architecture and weights variables, \textit{i.e.}, $D_i\left(\beta,w_{D_i}\right)$. $N$ individual discriminators are combined with the same generator $G^*$ to form $N$ GAN models. The networks are trained using a weight-sharing and many-to-one adversarial training strategy, and the top $N/2$ individuals are selected through the proposed fitness function $F_D$. We employ crossover \& mutation on the $N/2$ retained individuals to generate a new set of $N/2$ offspring, thus forming a new population of $N$ offspring. The optimizing object of the evolution generator architecture search can be formulated as:
	\begin{gather}
		\label{searchD}
		\beta^{*}=\arg \min _{\beta_{i}}\left\{F_{D}\left(\beta_{i} \mid w_{D_{i}}^{*}, w_{G^*}^{*}, \alpha^*\right), i \in\{1, \ldots, N\}\right\},
	\end{gather}
	\begin{gather}
		\label{trainD}
		\begin{array}{ll} 
			\begin{aligned}
				\text { s.t. } w_{G^*}^{*}=&\arg \min _{w_{G^*}} \sum_{i=1}^{N} \mathbb{E}_{z \sim p(z)}\left[-D_i\left(a, G^*(a, z)\right)\right]{\color{blue},} \\
				w_{D_i}^{*}=&\arg \max _{w_{D_i}} \mathbb{E}_{x \sim p_{\text {data }}(x)}[D_i(a, x)] \\
				&-\mathbb{E}_{z \sim p(z)}\left[D_i\left(a, G_{i}(a, z)\right)\right],
			\end{aligned}
		\end{array}
	\end{gather}
	where Eq.  \ref{searchD} corresponds to evaluation and selection, and Eq.  \ref{trainD} corresponds to the training of the generator and discriminator populations..

	\noindent \textbf{Many-to-One Adversarial Training}
	The training process of the discriminator also uses the weight-sharing and many-to-one strategy. Compared to the evolution generator architecture search that uses a fixed discriminator, we fix the architecture of the generator for the evolution discriminator architecture search, \textit{i.e.}, $G^*(\alpha, w_{G^*})$. In each round, we produce $N$ candidate discriminators to form the discriminator population $P_D$, where all candidate generators share the weights $W_D$ of the Large\_Discriminator (\textit{i.e.}, $D_L$). Each candidate $D_i$ is parameterized by its architecture and weight variables, \textit{i.e.}, $D_i(\beta_i, w_{D_i})$.
	
	We then train the $N$ GANs composed of $N$ discriminators and one fixed generator using the many-to-one strategy, updating the parameters of discriminators and generator using Eq. \ref{trainD}. These $N$ GANs models share the same discriminator that is trained for multiple epochs in each round. For each training batch, we uniformly sample a discriminator from the $N$ candidate discriminators and optimize it with the fixed generator. This training process is equivalent to activating and training the parameters $w_{D_i}$ corresponding to the $D_i$ architecture within $D_L$, as well as the parameters $w_{G^*}$ for the generator $G^*$.
	
	\noindent \textbf{Evaluation \& Selection:}
	After adversarial training, we introduce the fitness function $F_D$ to evaluate the quality of each discriminator. $F_D$ consists of two fitness components, namely quality fitness $F_{D^q}$ and architecture complexity $F_{D^c}$. The existing generative ZSL networks are based on Wasserstein GAN \cite{Arjovsky2017WassersteinGA}, in which the discriminator takes the unconstrained continul value as output compared to the real samples. As such, we can calculate the distances between the generated features and the real feature samples based on the discriminator, which can be applied to measure the quality for evaluation and selection:
	\begin{gather}
		\label{eq:F_Dq}
		\mathcal{F}_{D^q}=-\mathbb{E}_{x \sim p_{data}} D_i\left(a, x\right) + \mathbb{E}_{z \sim p(z)} D_i\left(a, G^*(a, z)\right).
	\end{gather}
	A smaller $F_{D^q}$ value indicates a better ability of the discriminator to distinguish the real and generated samples. To ensure model efficiency, we also take the architecture complexity $F_{D^c}$ for model evaluation:
	\begin{gather}
		\label{eq:F_Dc}
		\mathcal{F}_{D^c}= C_{E_{nn}}(D_i)/C_{E_{all}}(D_i),
	\end{gather}
	As such, the fitness function $F_D$ can be formulated as:
	\begin{gather}
		\label{eq:F_D}
		\mathcal{F}_{D}=\mathcal{F}_{D^q} - \lambda_D\times \mathcal{F}_{D^c}
	\end{gather}
	$\lambda_D > 0$ is used to balance the architecture complexity of the discriminator. Finally, based on $F_D$, we select the top $N/2$ individuals $\{G_1,\cdots, G_{N/2}\}$ to serve as the parents for the next generation by ascending the order:
	\begin{gather}
		\label{eq:select-D}
		\{D_1,\cdots,D_{N/2}\}\leftarrow sort_{min}\{\mathcal{F}_{D_i}\}_{i=1}^N.
	\end{gather}
	
	Then, we take the same crossover and mutation mechanism as the evolution generator architecture search to produce $N$ discriminator individuals for the evolution discriminator architecture search in the next round.

	\subsection{Zero-shot prediction}\label{sec3.2}
	After the optimization of the EGANS algorithm, we obtain the optimal generator (\textit{i.e.}, $G^*$) and discriminator (\textit{i.e.}, $D^*$) to form a GAN model, which is applied in the generative ZSL model. Specifically, we first take the seen class data $x\in \mathcal{D}_{tr}^s$ to train the generative ZSL model that is conditioned by the class semantic vectors $a$. Then we employ the pretrained conditional generator (\textit{i.e.}, $G^{**}$) to synthesize the features samples for unseen classes according to the $a^u$, \textit{i.e.}, $G^{**}(a^u,z)=\tilde{\mathcal{X}}$. Finally, we take $\bar{\mathcal{X}}=\{\tilde{\mathcal{X}} , \mathcal{X}^s\}$ (where $\mathcal{X}^s\in \mathcal{D}_{tr}^s$) to learn a supervised classifier (\textit{e.g.}, softmax) over $y\in\mathcal{Y}$, \textit{i.e.}, $f_{c z s l}: \tilde{\mathcal{X}}\rightarrow \mathcal{Y}^{u}$ or $f_{g z s l}: \bar{\mathcal{X}} \rightarrow \mathcal{Y}^{s} \cup \mathcal{Y}^{u}$. Once the classifier is trained, we use $x\in\{\mathcal{D}_te^s, \mathcal{D}_te^u\}$ to test the model for ZSL prediction.

	\section{Experiments and Evaluation}\label{sec4}
	In subsequent sections, we introduce the datasets, evaluation protocols, and implementation details. Furthermore, we provide a series of experiment analyses to verify our methods.
	
	\begin{figure*}[htbp]
		\begin{center}
			\includegraphics[width=0.5\linewidth]{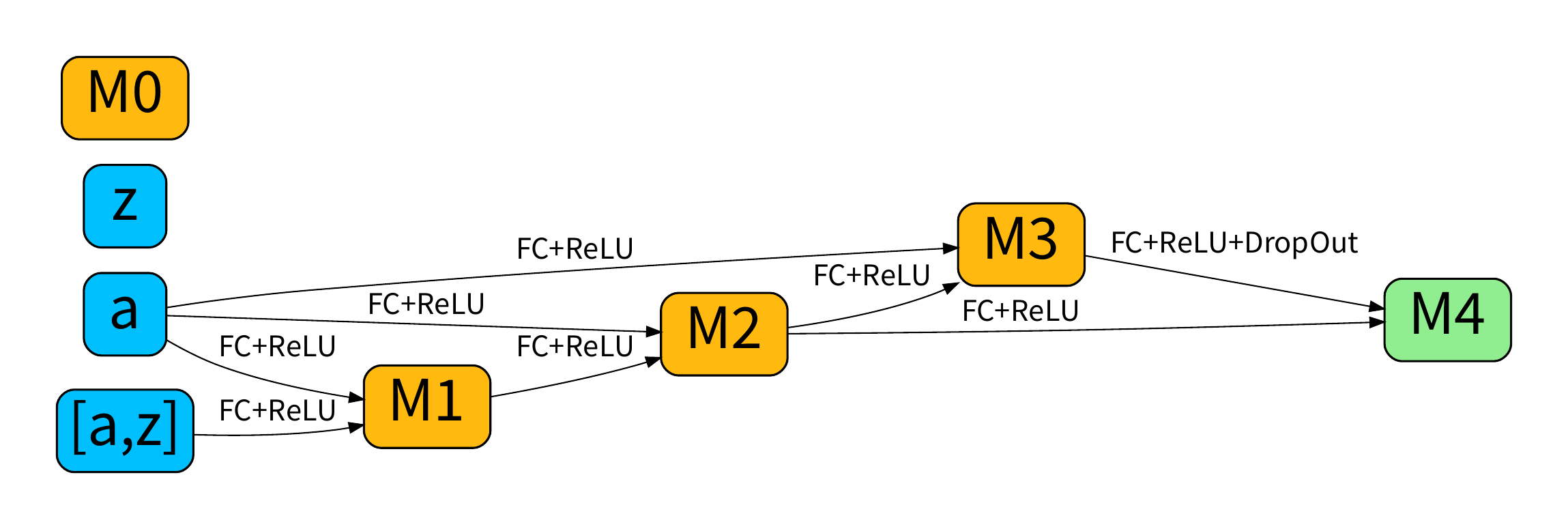}\hspace{5mm}
			\includegraphics[width=0.35\linewidth]{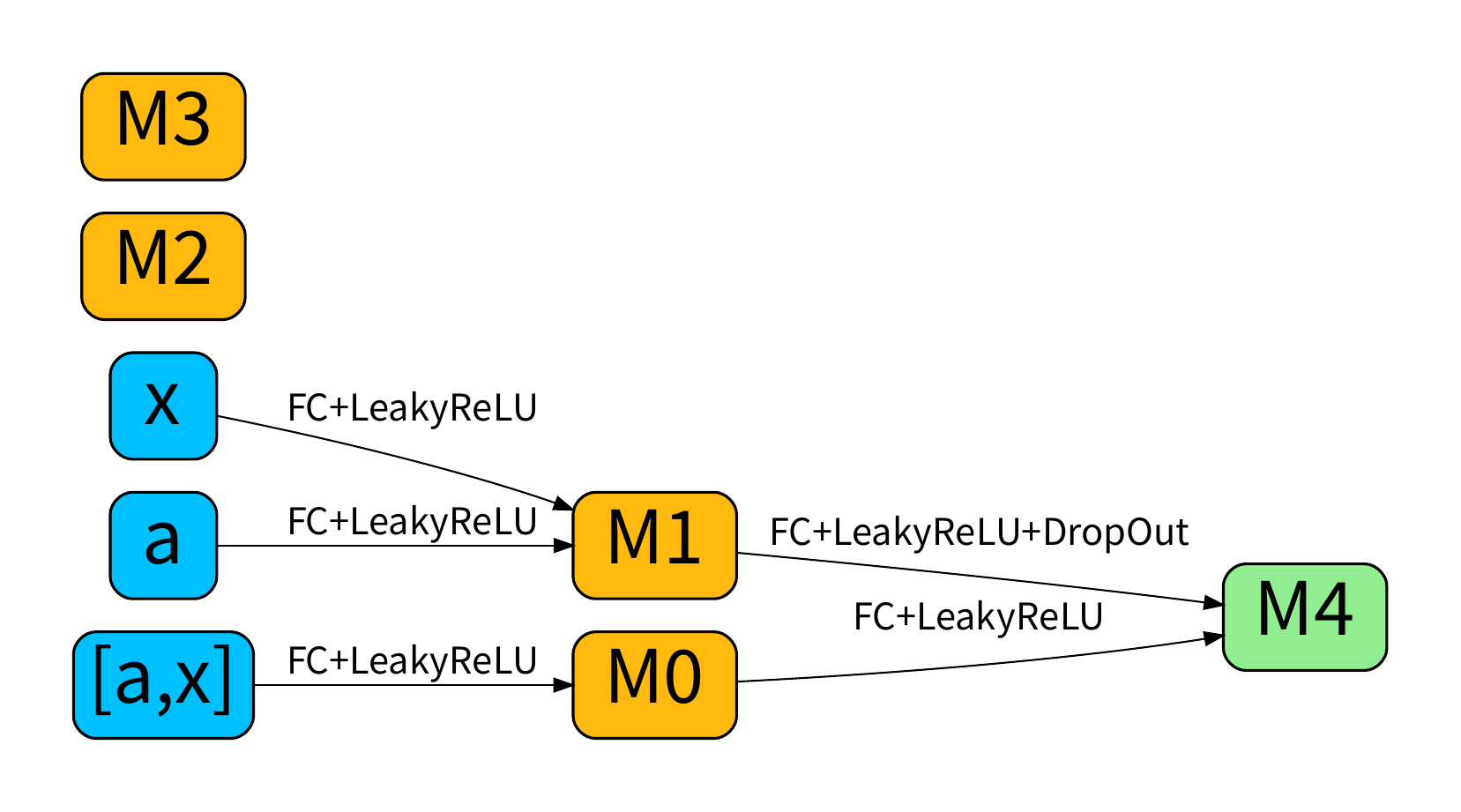}\\
			(a) CUB\_G \hspace{8cm} (b) CUB\_D \\
			\includegraphics[width=0.5\linewidth]{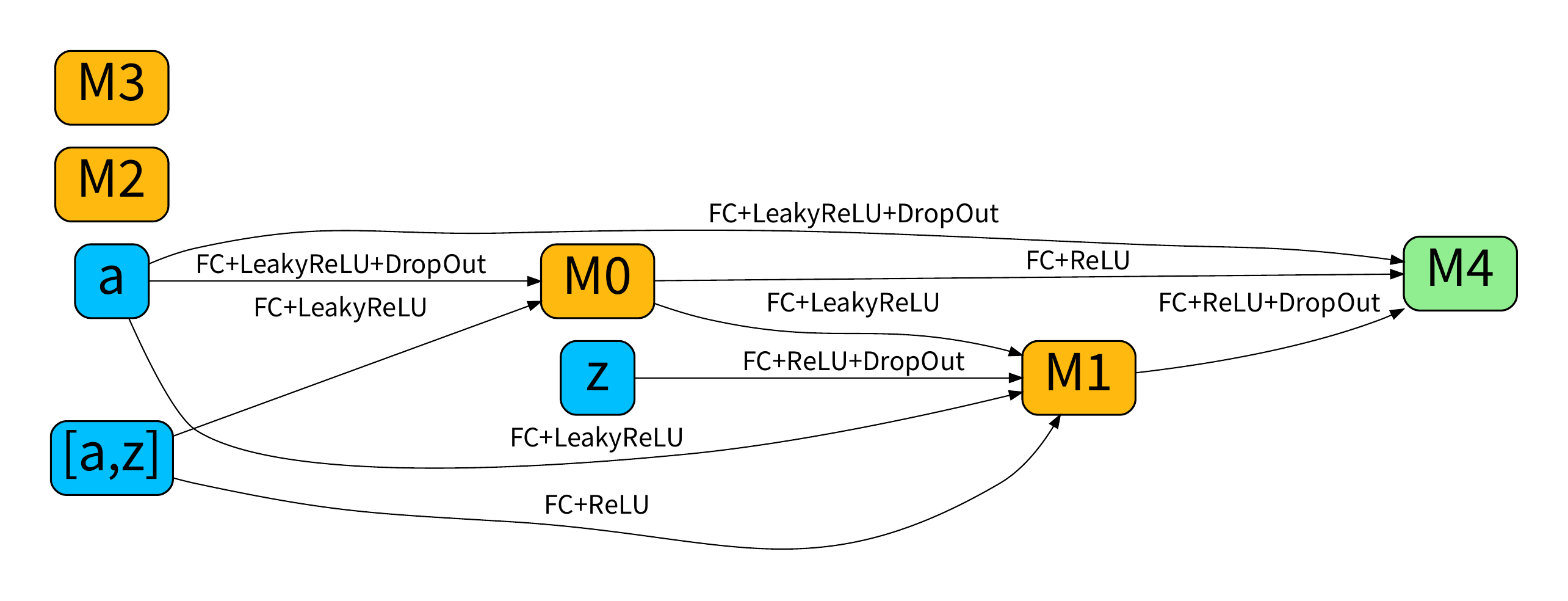}
			\includegraphics[width=0.45\linewidth]{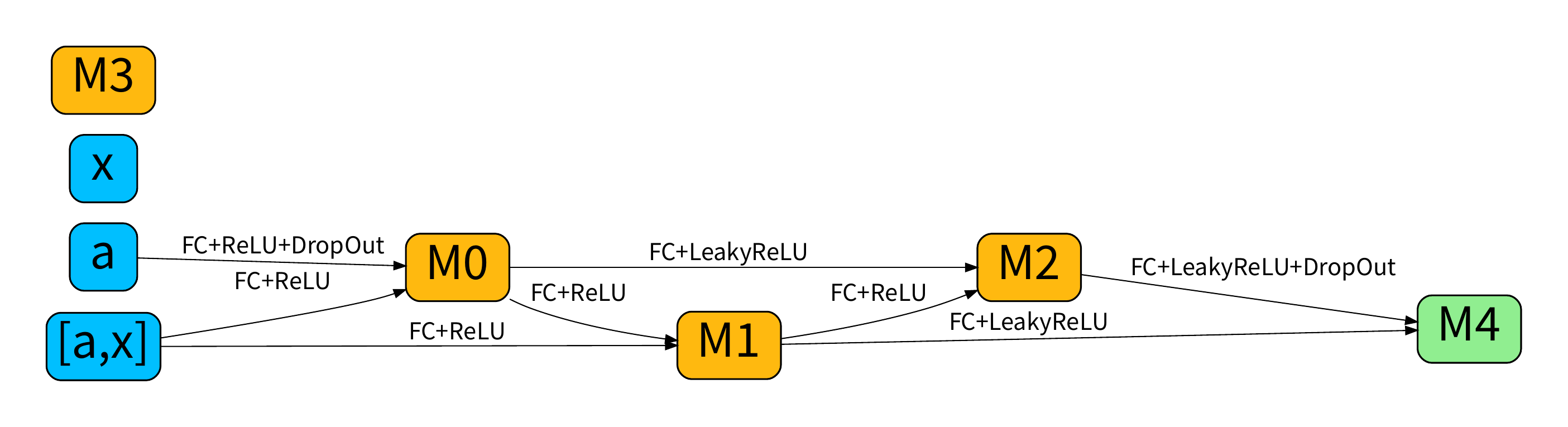}\\
			(c) SUN\_G \hspace{8cm} (d) SUN\_D \\
			\includegraphics[width=0.5\linewidth]{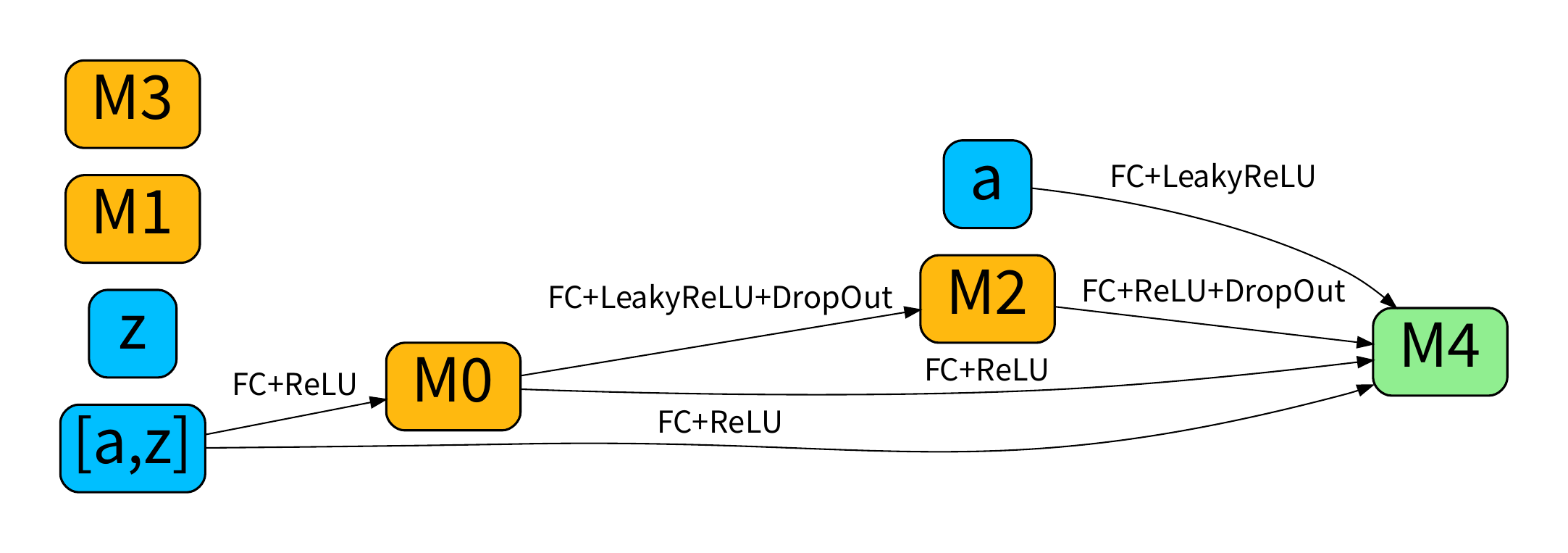}
			\includegraphics[width=0.45\linewidth]{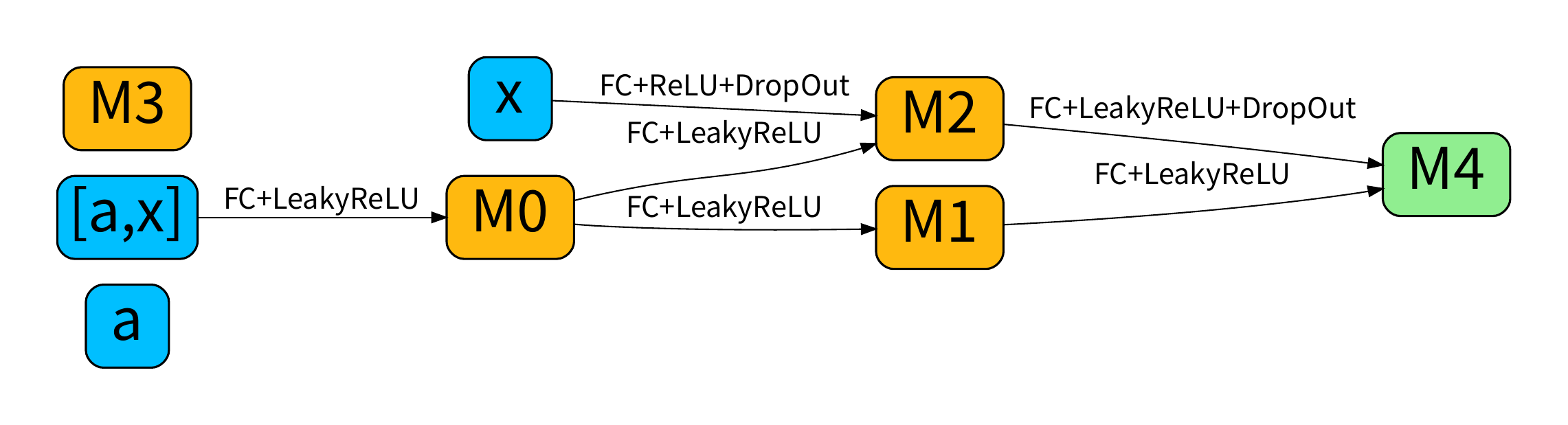}\\
			(e) FLO\_G \hspace{8cm} (f) FLO\_D \\
			\includegraphics[width=0.35\linewidth]{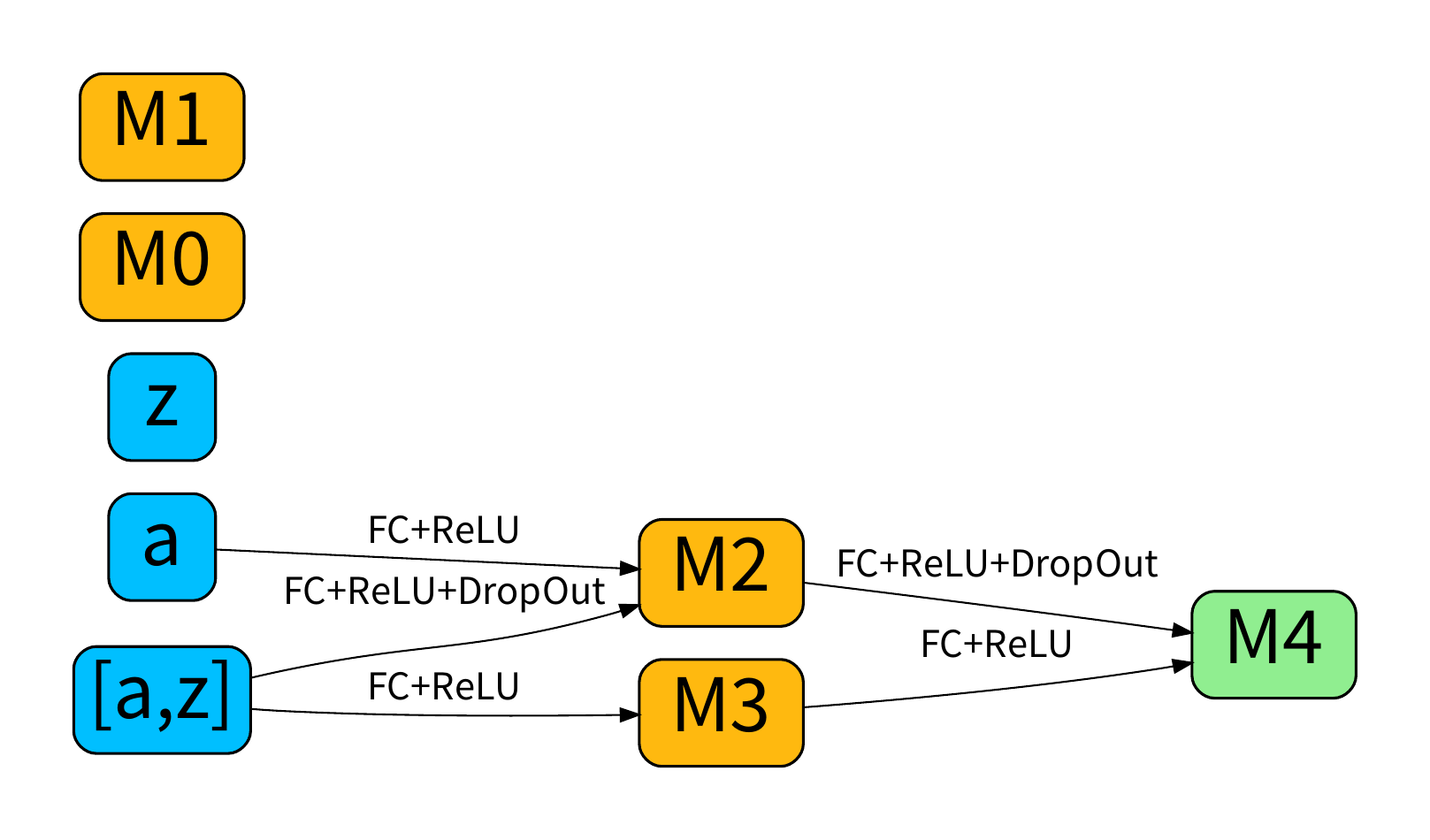} \hspace{15mm}
			\includegraphics[width=0.35\linewidth]{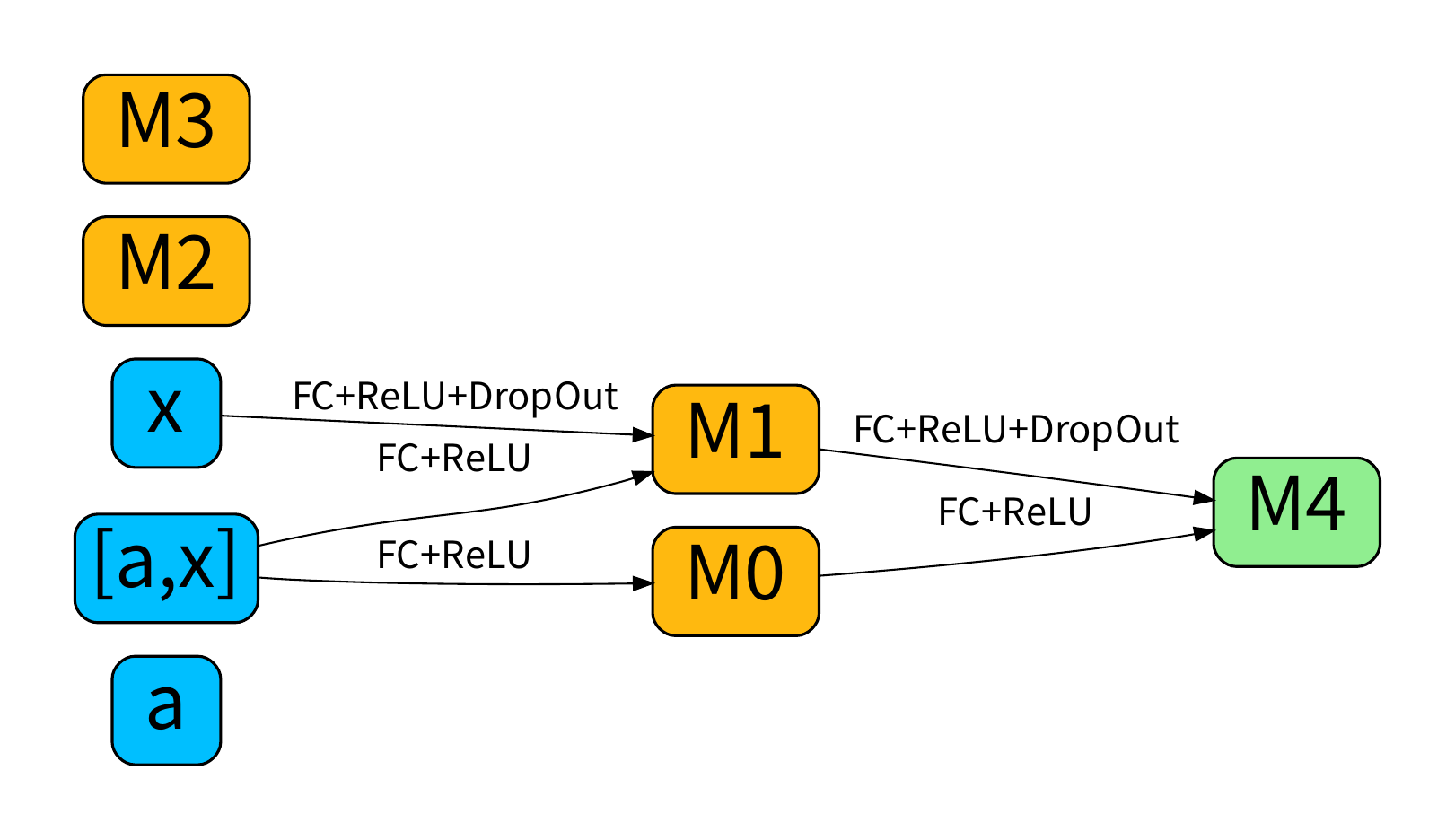}\\
			(g) AWA2\_G \hspace{8cm} (h) AWA2\_D \\
			\caption{The searched generator and discriminator architecture pairs on CUB (a and b), SUN (c and d), FLO (e and f), and AWA2 (g and h). The “a”, “z” and “x” in the input space denote the class semantic vector, noise vector, and visual feature representation, respectively, while “o” denotes the output node.}
			\label{fig:architecture}\vspace{-6mm}	
		\end{center}
	\end{figure*}
	
	\subsection{Datasets}\label{sec4.00}
	We evaluate our method on four widely-used benchmark datasets collected from various \textbf{scenarios}, \textit{i.e.}, CUB (Caltech UCSD Birds 200) \cite{Welinder2010CaltechUCSDB2}, SUN (SUN Attribute) \cite{Patterson2012SUNAD},  FLO (Oxford Flowers) \cite{Nilsback2008AutomatedFC}, and  AWA2 (Animals with Attributes 2) \cite{Xian2019ZeroShotLC}. CUB consists of 11,788 images of 200 \textbf{bird} classes with 312 attributes. SUN includes 14,340 images from 717 \textbf{scene} classes described by 102 attributes. FLO contains 8189 images of 102 \textbf{flower} classes with 1024 attributes. AWA2 includes 37,322 images from 50 \textbf{animal} classes captured by 85 attributes. Notably, CUB, SUN, and FLO  are fine-grained datasets, whereas AWA2 is the coarse-grained dataset. Following \cite{Xian2019ZeroShotLC}, we take the same dataset splits, which are summarized in Table \ref{Table:dataset}.
	\begin{table}[h]
		\centering
		\caption{Statistics of the CUB, SUN, FLO and AWA2 datasets, \textit{i.e.}, the dimensions of semantic vectors per class (denoted as |A|), seen/unseen class
			number (denoted as \textit{Seen/Unseen}), and total number of samples (denoted as \textit{Num}).} \label{Table:dataset}
		\setlength{\tabcolsep}{6mm}{
			\begin{tabular}{l|c|c|c}
				\hline       
				&  |A|  &      \textit{Seen/Unseen}       & \textit{Num} \\   
				\hline
				CUB \cite{Welinder2010CaltechUCSDB2}   & 312 &  150/50   & 11788  \\
				SUN \cite{Patterson2012SUNAD}  & 102 &  645/72   & 14340  \\
				FLO \cite{Nilsback2008AutomatedFC}  & 1024  &  82/20  & 8189   \\
				AWA2 \cite{Xian2019ZeroShotLC} & 85  &  40/10    & 37322  \\
				\hline
				
		\end{tabular}}
	\end{table}
	
	\begin{table*}[ht]
		\centering  
		\caption{Comparing with existing popular ZSL methods on four datasets in the GZSL setting. The first part and second part include embedding-based ZSL and generative ZSL methods, respectively.}
		\resizebox{\linewidth}{!}{\small
			\begin{tabular}{l|ccl|ccl|ccl|ccl}
				\hline
				\multirow{2}{*}{\textbf{Methods}} 
				&\multicolumn{3}{c|}{\textbf{CUB}}&\multicolumn{3}{c|}{\textbf{SUN}}&\multicolumn{3}{c}{\textbf{FLO}}&\multicolumn{3}{c}{\textbf{AWA2}}\\
				\cline{2-4}\cline{5-7}\cline{8-10}\cline{11-13}
				&\rm{U} & \rm{S} & \rm{H} &\rm{U} & \rm{S} & \rm{H} &\rm{U} & \rm{S} & \rm{H} &\rm{U} & \rm{S} & \rm{H}  \\
				\hline
				CMT~\cite{Socher2013ZeroShotLT}&7.2 &49.8 &12.6&  8.1& 21.8& 11.8&--&--&--&0.5&90.0& 1.0\\
				ALE~\cite{Akata2016LabelEmbeddingFI}& 23.7&62.8& 34.4& 21.8&33.1& 26.3&--&--&--&14.0& 81.8& 23.9\\
				ESZSL~\cite{RomeraParedes2015AnES}& 12.6 &63.8& 21.0& 11.0& 27.9& 15.8&--&--&--& 5.9 &77.8&11.0\\
				DCN~\cite{Liu2018GeneralizedZL}&28.4&60.7&38.7&25.5&37.0&30.2&--&--&--&--&--&--\\
				\hline
				GAZSL~\cite{Zhu2017AGA}& 22.7&63.2& 33.4& 21.2&38.4& 27.3& 30.1& 81.9&44.0&22.5 &80.0 &35.1\\
				GZSL-OD~\cite{Mandal2019OutOfDistributionDF}&35.8& 46.2& 40.3& 26.6&32.5& 29.3&55.2&57.8& 56.4&47.9& 69.1& 56.6\\
				CLSWGAN~\cite{Xian2018FeatureGN}&43.7&57.7&49.7&42.6&36.6&39.4&57.9&74.9&65.3&57.9&61.4&59.6\\
				CLSWGAN~\cite{Xian2018FeatureGN}+\textbf{EGANS}&\textbf{47.8}&\textbf{57.4}&\textbf{52.2}\textbf{$^{\color{blue}\uparrow\text{\textbf{2.5}}}$}& \textbf{46.9}&\textbf{35.8}&\textbf{40.6}$^{\color{blue}\uparrow\text{\textbf{1.2}}}$&\textbf{57.9}&\textbf{83.9}&\textbf{68.5}$^{\color{blue}\uparrow\text{\textbf{3.2}}}$&\textbf{58.1}&\textbf{69.1}&\textbf{63.1}$^{\color{blue}\uparrow\text{\textbf{3.5}}}$\\
				LisGAN~\cite{Li2019LeveragingTI}&46.5&57.9&	51.6&42.9&37.8&40.2&57.7&83.8&	68.3&52.6&76.3&62.3\\
				LisGAN~\cite{Li2019LeveragingTI}+\textbf{EGANS}&\textbf{47.8}&\textbf{59.2}&\textbf{52.9}\textbf{$^{\color{blue}\uparrow\text{\textbf{0.5}}}$}& \textbf{44.2}&\textbf{37.4}&\textbf{40.5}$^{\color{blue}\uparrow\text{\textbf{0.3}}}$&\textbf{60.6}&\textbf{87.3}&\textbf{71.5}$^{\color{blue}\uparrow\text{\textbf{3.2}}}$&\textbf{53.9}&\textbf{81.8}&\textbf{65.0}$^{\color{blue}\uparrow\text{\textbf{2.7}}}$\\
				\hline
		\end{tabular} }
		\label{table:SOTA-GZSL} 
	\end{table*}

	\begin{table}[ht]
		\centering  
		\vspace{-2mm}
		\caption{Comparing with existing popular ZSL methods on four datasets in the CZSL setting. The first part and second part include embedding-based ZSL and generative ZSL methods, respectively.}
		\resizebox{\linewidth}{!}{\small
			\begin{tabular}{l|l|l|l|l}
				\hline
				\multirow{2}{*}{\textbf{Methods}} 
				&\multicolumn{1}{c|}{\textbf{CUB}}&\multicolumn{1}{c|}{\textbf{SUN}}&\multicolumn{1}{c}{\textbf{FLO}}&\multicolumn{1}{c}{\textbf{AWA2}}\\
				\cline{2-5}
				&\rm{acc} &\rm{acc} & \rm{acc} & \rm{acc}\\
				\hline
				CMT~\cite{Socher2013ZeroShotLT}&34.6& 39.9&--&37.9\\
				ALE~\cite{Akata2016LabelEmbeddingFI}&54.9& 58.1&--&62.5\\
				ESZSL~\cite{RomeraParedes2015AnES}&53.9& 54.5&--&58.6\\
				DCN~\cite{Liu2018GeneralizedZL}&56.2& 61.8&--&65.2\\
				\hline
				GAZSL~\cite{Zhu2017AGA}&55.7& 60.7&60.6&--\\
				GZSL-OD~\cite{Mandal2019OutOfDistributionDF}&58.7&59.1&63.7&--\\
				CADA-VAE~\cite{Schnfeld2019GeneralizedZA} &59.8&61.7&--&63.0\\
				\cline{2-5}
				CLSWGAN~\cite{Xian2018FeatureGN}& 57.3&58.9& 66.0& 68.2\\
				CLSWGAN~\cite{Xian2018FeatureGN}+\textbf{EGANS}& \textbf{59.2}$^{\color{blue}\uparrow\text{\textbf{1.9}}}$&\textbf{61.5}$^{\color{blue}\uparrow\text{\textbf{2.6}}}$& \textbf{67.3}$^{\color{blue}\uparrow\text{\textbf{1.3}}}$& \textbf{69.9}$^{\color{blue}\uparrow\text{\textbf{1.7}}}$\\
				\cline{2-5}
				LisGAN~\cite{Li2019LeveragingTI}& 58.8&61.7& 68.3& 70.1\\
				LisGAN~\cite{Li2019LeveragingTI}+\textbf{EGANS}& \textbf{60.2}$^{\color{blue}\uparrow\text{\textbf{1.4}}}$&\textbf{62.8}$^{\color{blue}\uparrow\text{\textbf{1.1}}}$& \textbf{69.5}$^{\color{blue}\uparrow\text{\textbf{1.2}}}$& \textbf{70.6}$^{\color{blue}\uparrow\text{\textbf{0.5}}}$\\
				\hline
		\end{tabular} }
		\label{table:SOTA-CZSL} 
	\end{table}

	\subsection{Evaluation Protocols}\label{sec4.0}
	Following \cite{Xian2019ZeroShotLC}, we take the top-1 accuracy for evaluation both in the CZSL and GZSL settings. In the CZSL setting, we just calculate the accuracy of unseen, as the test samples are only from unseen classes, denoted as $\bm{acc}$. In the GZSL setting, the test samples include seen and unseen classes. Thus, we calculate the accuracy of the test samples from both seen classes (denoted as $\bm{S}$) and unseen classes (denoted as $\bm{U}$). Typically, the harmonic mean (defined as $\bm{H =(2 \times S \times U) /(S+U)}$) of seen and unseen classes is the most important metric for evaluating the performance of GZSL.

	\subsection{Implementation Details}\label{sec4.1}

	We take the ResNet101 pretrained on ImageNet with an input size of $224\times224$ to extract the visual features $x \in \mathbb{R}^{2048}$ without fine-tuning. We set up a warm-up strategy before the evolution search to ensure fair competition for all candidate subnets. Specifically, all candidate operations in the search space are activated uniformly and trained equally with five warm-up epochs. After that, we randomly sample $N$ subnets to form the first round of population for both the evolution generator/discriminator architecture search. We take the Adam optimizer with hyperparameters (betas=(0.5, 0.999)) to optimize the network. The learning rate and batch size are set to 0.0001 and 64, respectively. The architecture complexity weights $\lambda_{G}$ and $\lambda_{D}$ are set to 0.1 and 1.5.  All experiments are performed on a single NVIDIA 1080 graphic card with 11GB memory. We use PyTorch\footnote{\url{https://pytorch.org/}} to implement all experiments.

	\subsection{Experiment 1: Searched GAN Architecture}\label{sec4.2}
	As shown in Fig. \ref{fig:architecture}, we presented the discovered generator and discriminator pairs for each dataset. From these results, we can conclude the following observations: i) For each paired generator and discriminator, the model complexity of the generator is larger than the corresponding discriminator. The reason is that we mainly utilize the powerful generator (functioned as semantic$\rightarrow$visual mapping) to synthesize visual features for data augmentation in ZSL. ii) The input node of all the searched architecture involves concatenated semantic and visual information, as the generator and discriminator conditioned by the class semantic vector help the learning of the generator and discriminator with class information supervision, enabling the synthesize visual features are class-relevant. This is consistent with the manually designed architectures in the generative ZSL models, \textit{e.g.}, CLSWGAN \cite{Xian2018FeatureGN}. iii) The searched generators on the fine-grained datasets (\textit{e.g.}, CUB, SUN, and FLO) are more complex than the ones searched on the coarse-grained dataset (\textit{e.g.}, AWA2). This phenomenon shows that the fine-grained datasets require synthesizing the visual features captured by more detailed appearances, while the coarse-grained datasets just require some simple feature representations (\textit{e.g.}, the object size) to distinguish various classes. This demonstrates that the fixed architectures cannot well adapt to various datasets/scenarios, and thus automatically designing the GAN architecture with stable optimization customized for each specific ZSL task is very necessary and has great potential.

	\subsection{Experiment 2: Architecture Evaluation}\label{sec4.3}
	In this section, we qualitatively and quantitatively analyze the ZSL performance of GAN architecture searched by our EGANS. Here, we take f-CLSWGAN \cite{Xian2018FeatureGN} and LisGAN \cite{Li2019LeveragingTI} as baselines for comparison and discussion.

	\subsubsection{Quantitative Evaluation}\label{sec4.3.1}
	Our EGANS can be entailed into existing generative ZSL methods (\textit{e.g.}, CLSWGAN~\cite{Xian2018FeatureGN} and LisGAN~\cite{Li2019LeveragingTI}). We compare it with other popular methods (\textit{i.e.}, embedding-based methods, and generative methods) both in CZSL and GZSL settings to demonstrate its effectiveness and advantages.

	Table \ref{table:SOTA-GZSL} shows the results of different methods in the GZSL setting, where both seen and unseen classes are available for prediction. Compared to the embedding-based methods, which achieve good performance on seen classes but fail on unseen classes, our EGANS enable generative methods (\textit{e.g.}, CLSWGAN~\cite{Xian2018FeatureGN} and LisGAN~\cite{Li2019LeveragingTI}) to achieve balanced and high unseen and seen accuracy. For example, LisGAN+EGANS obtains 14.2\% and 10.3\% improvements of Harmonic mean on CUB and SUN over the best-compared method (e.g., DCN \cite{Liu2018GeneralizedZL}), respectively. This demonstrates the effectiveness of i)  generative models that are beneficial for ZSL via data augmentation; and ii) our EGANS finds the optimal GAN architecture for various datasets, enabling the generative ZSL methods to adapt to various scenarios with desirable generalization.
	
	We also compare our method with the popular ZSL methods in the CZSL setting. As shown in Table \ref{table:SOTA-CZSL}, our EGANS significantly improves the performance of CLSWGAN~\cite{Xian2018FeatureGN} (by 1.9\%, 2.6\%, 1.3\% and 1.7\% on CUB, SUN, FLO and AWA2, respectively) and LisGANLisGAN~\cite{Li2019LeveragingTI} (by 1.4\%, 1.1\%, 1.2\% and 0.5\% on CUB, SUN, FLO  and AWA2, respectively). These results show that our EGANS can enhance the generalization by automatically searching optimal GAN models. Compared to the embedding-based methods, our EGANS helps existing generative ZSL methods gain state-of-the-art results with 60.2\%, 62.8\%, 69.5\% and 70.6\% on CUB, SUN, FLO and AWA2, respectively.  This further shows the advantages of EGANS for the generative model in the zero-shot learning task.

	\begin{figure}[htbp]
		\begin{center}
			\includegraphics[width=0.98\linewidth]{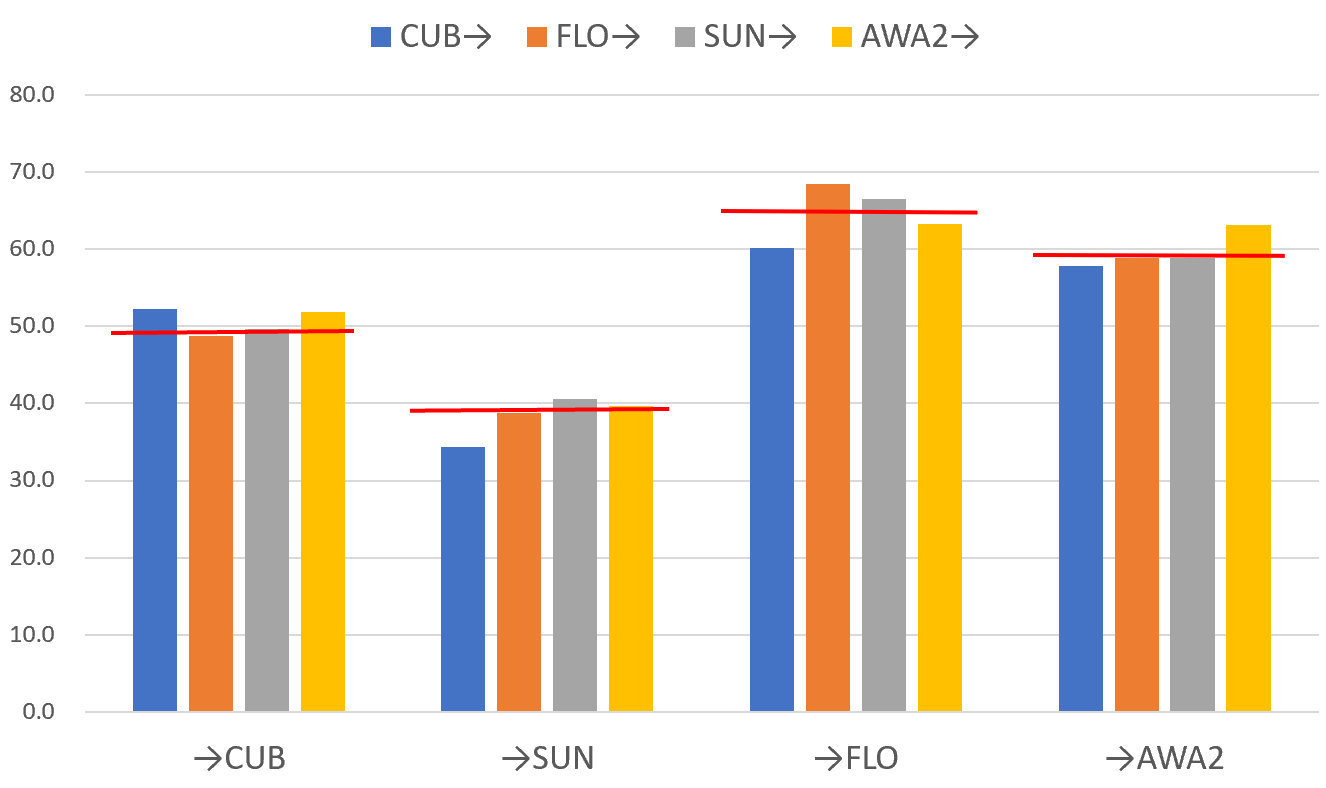}\\
			\caption{The generalization of searched GAN architectures on various datasets in terms of harmonic mean in the GZSL setting, where the red line indicates the performance of CLSWGAN based on the manually-designed fixed GAN architecture that is used in all datasets.}
			\label{fig:generalization}
		\end{center}
	\end{figure}

	\begin{figure*}[htbp]
		\begin{center}
			\includegraphics[width=0.98\linewidth]{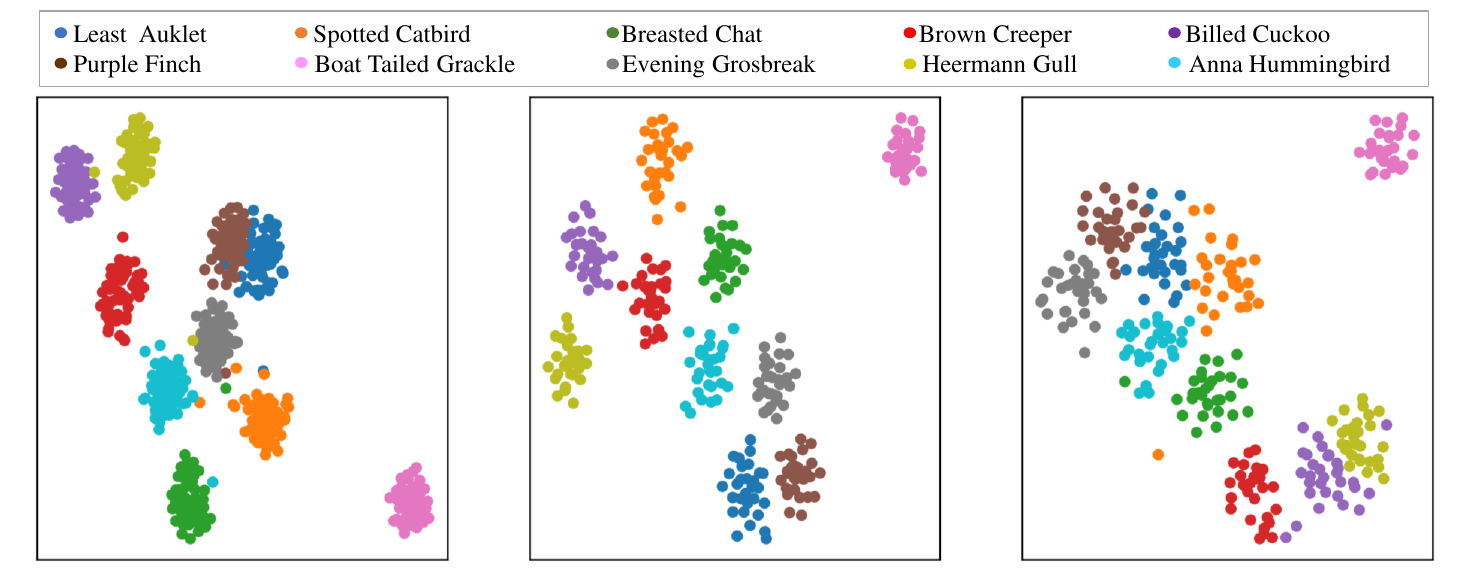}\\
			\hspace{1.1cm}(a) Real visual features \hspace{1.6cm} (b) Synthesized by CLSWGAN  \hspace{1cm} (c) Synthesized by CLSWGAN+EGANS\\
			\caption{The searched generator and discriminator architecture pairs on CUB (a and b), SUN (c and d), FLO (e and f), and AWA2 (g and h). The “a”, “z” and “f” in the input space denote the class semantic vector, noise vector and feature vector, respectively, while “o” denotes the output node. The 10 colors denote 10 different classes randomly selected from CUB.}
			\label{fig:t-sne}
		\end{center}
	\end{figure*}
	
	\begin{table*}[ht]
		\centering  
		\setlength{\aboverulesep}{0mm} 
		\setlength{\belowrulesep}{0mm}
		\caption{Ablation studies for different components of EGANS on four datasets.}
		\resizebox{1.0\linewidth}{!}{
			\begin{tabular}{r|c|ccc|c|ccc|c|ccc|c|ccc}
				\toprule
				\multirow{3}{*}{\textbf{Methods}} 
				&\multicolumn{4}{c|}{\textbf{CUB}}&\multicolumn{4}{c|}{\textbf{SUN}}&\multicolumn{4}{c}{\textbf{FLO}}&\multicolumn{4}{c}{\textbf{AWA2}}\\
				\cline{2-5}\cline{6-9}\cline{10-13}\cline{14-17}
				&\multicolumn{1}{c|}{CZSL}&\multicolumn{3}{c|}{GZSL}&\multicolumn{1}{c|}{CZSL}&\multicolumn{3}{c|}{GZSL}&\multicolumn{1}{c|}{CZSL}&\multicolumn{3}{c}{GZSL}\\
				\cline{2-5}\cline{6-9}\cline{10-13}\cline{14-17}
				\textbf{} 
				&\rm{acc}&\rm{U} & \rm{S} & \rm{H} &\rm{acc}&\rm{U} & \rm{S} & \rm{H} &\rm{acc}&\rm{U} & \rm{S} & \rm{H} &\rm{acc}&\rm{U} & \rm{S} & \rm{H}  \\
				\hline 
				EGANS (Fixed G) & 57.9& 45.7&58.8& 51.4&56.1&41.9&35.5&38.4&65.8& 58.9& 77.1&66.8&67.7&57.3&68.9&62.5\\
				EGANS (Fixed D) & 58.3& 50.0&53.2& 51.6&59.0&41.5&38.1&39.7&65.1& 60.5& 77.4&67.9&68.0&58.2&68.2&62.8\\
				EGANS (full) & 59.2& 47.8&57.4& 52.2&61.5&46.9&35.8&40.6&67.3& 57.9& 83.9&68.5&68.2&58.1&69.1&63.1\\
				\bottomrule	
		\end{tabular} }
		\label{table:ablation}
	\end{table*}
	
	\subsubsection{Qualitative Evaluation}\label{sec4.3.2}
	To intuitively verify that the searched architectures of EGANS are able to learn the data distribution of unseen classes, we take 2-D t-SNE to visualize the real and synthesized visual features learned by CLSWGAN \cite{Xian2018FeatureGN} and CLSWGAN+EGANS on 10 classes randomly selected from CUB. If the synthesized visual features are diverse and share similar relationships between various classes with the real visual features, we argue that the GAN model can capture the data distribution of unseen classes for data augmentation in the ZSL task. As shown in Fig. \ref{fig:t-sne}, we find that i) the synthesized visual features learned by CLSWGAN cannot well keep the true relationships between various classes compared to the real visual features, while our CLSWGAN+EGANS learn the accurate data distribution as same as the real data. For example, the visual features of \textit{Spotted Catbird} should be closer to the ones of \textit{Anna Hummingbird} than \textit{Brown Creeper}. ii) the visual features of each class produced by CLSWGAN+EGANS are diverse, while the visual features of each class synthesized by CLSWGAN locate in a small range caused by the model collapse of GAN. These observations demonstrate that EGANS discover optimal GAN architecture with good generalization and stability, and enable generative models to synthesize reliable visual features for data augmentation in ZSL.

	\subsubsection{Generalization Evaluation}\label{sec4.3.3}
	To evaluate the generalization of the searched architectures, we apply the GAN architecture searched on one dataset to the other datasets (\textit{e.g.}, CUB$\rightarrow$SUN, CUB$\rightarrow$FLO, and CUB$\rightarrow$AWA2). As shown in Fig. \ref{fig:generalization}, we present the results of harmonic mean in the GZSL setting. Results show that i) the transferred architectures suffer from serious performance degradation compared to their own architectures searched by EGANS; and ii) the results of baseline (\textit{i.e.}, CLSWGAN \cite{Xian2018FeatureGN}) are poorer than the data-specific searched architecture by EGANS. These results demonstrate that the searched GAN architectures by EGANS are data-dependent and effective, and thus the mutually-designed GAN architecture cannot well adapt to various datasets/scenarios.

	\begin{figure*}[htbp]
		\begin{center}
			\includegraphics[width=0.98\linewidth]{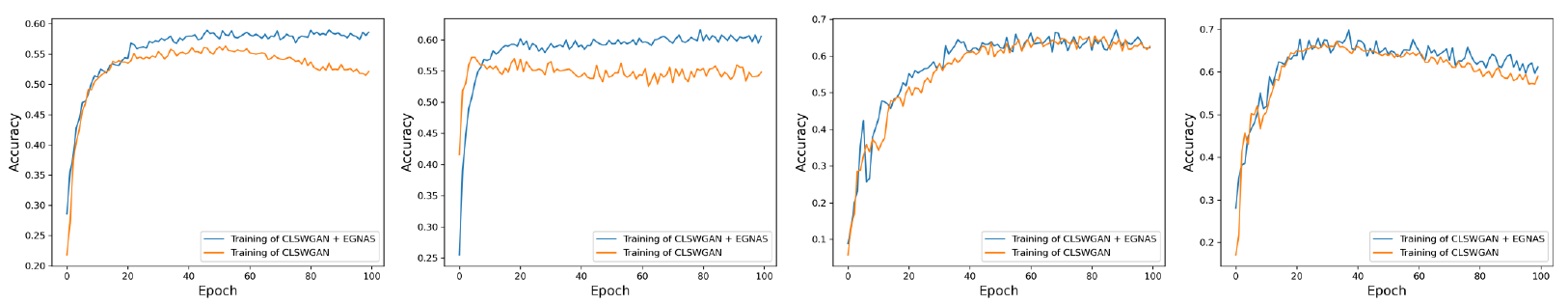}\\
			\hspace{0.8cm}(a) CUB \hspace{3cm} (b) SUN  \hspace{3cm} (c) FLO  \hspace{3cm} (c) AWA2\\
			\caption{Top-1 accuracy in the CZSL setting over four benchmark datasets during the training of CLSWGAN and our CLSWAN+EGANS.}
			\label{fig:trianing-efficiency}
		\end{center}
	\end{figure*}
	\begin{table*}[ht]
		\centering  
		\setlength{\aboverulesep}{0mm} 
		\setlength{\belowrulesep}{0mm}
		\caption{Time consuming for the evolution architecture search of generator and discriminator in our EGANS on CUB with one Nvidia 1080Ti GPU.}
		\resizebox{0.9\linewidth}{!}{
			\begin{tabular}{r|cc}
				\toprule
				Method  & EGANS(Search Generator) & EGANS(Search Discriminator) \\
				\hline
				Time Consuming (mins) &  32.8& 107.6\\
				\bottomrule	
		\end{tabular} }
		\label{table:comput-cost}
	\end{table*}

	\subsection{Experiment 3: Ablation Study}\label{sec4.4}
	We conduct ablation studies to evaluate the effectiveness of our EGANS in terms of the evolutionary generator search (denoted as EGANS (Fixed D)) and evolutionary discriminator search (denoted as EGANS (Fixed G)). As shown in Table \ref{table:ablation}, EGANS performs poorer than its full model only using evolutionary discriminator search cooperated with fixed G, \textit{i.e.}, the acc/H in CZSL/GZSL drop by 1.3\%/0.8\%, 5.4\%/2.2\%, 1.5\%/1.7\% and 0.5\%/0.6\% on CUB, SUN and FLO, respectively. Analogously, the performance of EGANS also decreases on all datasets when only using evolutionary discriminator search cooperated with fixed G. These results intuitively show that it is necessary to simultaneously search the architectures for G and D to construct an optimal GAN model, enabling to synthesize reliable visual features of unseen classes for data augmentation in ZSL. Furthermore, we find that searching a generator is more important than searching a discriminator, as we mainly take the generator to synthesize the visual features during ZSL classification.

	\subsection{Experiment 4: Training Efficiency Evaluation}\label{sec4.5}
	To show the training efficiency of the GAN model, we report the accuracy in the CZSL setting along with the training processes of CLSWGAN\cite{Xian2018FeatureGN}  and CLSWGAN+EGANS on four datasets. As shown in Fig. \ref{fig:trianing-efficiency}, results show that our CLSWGAN+EGANS consistently achieves better performance on all datasets than CLSWGAN. Meanwhile, we also observe that our CLSWGAN+EGANS  preserves comparable convergence speed over CLSWGAN. This is because the cooperative evolution for the generator and discriminator helps EGANS search optimal architectures for them, and thus the searched GAN model is more stable and avoids the mode collapse issue.
	
	Furthermore, we also provide the specific computational cost of the evolution architecture search of the generator and discriminator in our EGANS. Results are shown in Table \ref{table:comput-cost}. Overall, EGANS searches the network architectures of the generator and discriminator quickly. For example, the search times for the generator and discriminator in one Nvidia 1080Ti GPU are 32.8 and 107.6 minutes on CUB, respectively. This is because the well-designed search space and search strategy encourage EGANS to be optimized effectively.
	
	\begin{figure}[h]
		\begin{center}
			\hspace{-5mm}\includegraphics[width=4.3cm,height=3.2cm]{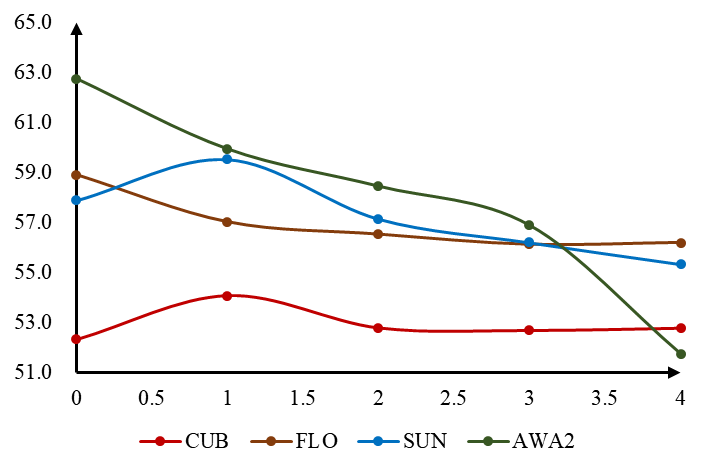}
			\includegraphics[width=4.3cm,height=3.2cm]{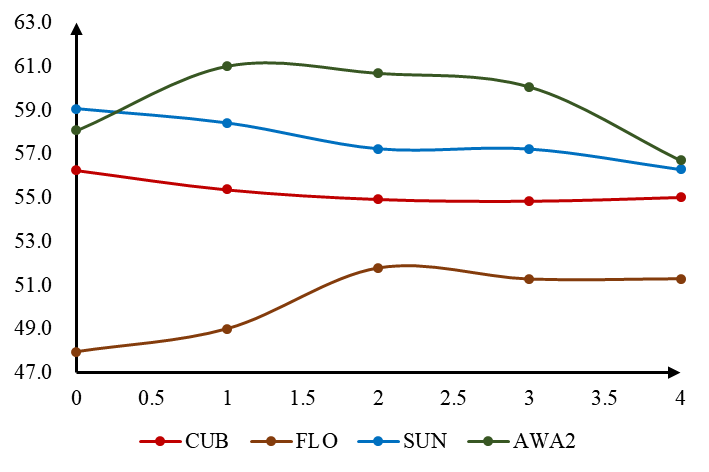}\\
			(a) $\lambda_{G}$ \hspace{3.8cm} (b) $\lambda_{D}$
			\caption{The effectiveness of coefficients $\lambda_{G}$ and $\lambda_{D}$ in the fitness functions of generator and discriminator, respectively.}
			\label{fig:para}
		\end{center}
	\end{figure}
	
	\subsection{Experiment 5: Hyper-Parameter Analysis}\label{sec4.6}
		We investigate the impact of hyper-parameter in our EGANS with respective to the coefficients $\lambda_{G}$ (in Eq. \ref{eq:F_G}) and $\lambda_{D}$ (in Eq. \ref{eq:F_D}). As shown in Fig. \ref{fig:para}, we provide the harmonic mean with respect to various value settings of $\lambda_{G}$ and $\lambda_{D}$ on four datasets. During searching generator, we should set a small $\lambda_{G}$ to control the complexity of generator, \textit{i.e.}, $\lambda_{G}\in [0,1]$. In contrast, we should set $\lambda_{D}$ to a large value (\textit{e.g.}, $\lambda_{D}\in[1,3]$), which encourages the powerful discriminator to provide the significant signal for the evolution of generator.

	\section{Discussion}\label{sec5}
	
	We would like to have more discussion here about the advantages and limitations of the proposed EGANS. First, we analyze why and how our method can automatically discover the optimal GAN architecture for the ZSL task.
	\begin{itemize}
		
		\item  Compared to the existing generative ZSL methods that apply the manually designed GAN model to various datasets/scenarios (\textit{e.g.}, birds, scene, flowers, and animals), we take the NAS technique to automatically search the optimal GAN models that significantly adapt to various datasets in ZSL task. For example, the search generator architectures are more complex on fine-grained datasets (\textit{e.g.}, CUN and SUN) than on coarse-grained datasets (\textit{e.g.}, AWA2), as the generators are required to synthesize more powerful feature representations to capture the detail appearance on fine-grained datasets. Experiments in Section \ref{sec4.3.1} and \ref{sec4.3.3} intuitively support these claims.

		\item  As discussed in \cite{Gong2019AutoGANNA, Wang2019AGANTA, Gao2019AdversarialNASAN}, the search complexity and the instability of GAN training result in difficult optimization for search GAN architecture. Accordingly, we introduce the cooperative dual evolution for the generator and discriminator, enabling the optimization to be more stable during an adversarial evolution search. As such, our EGANS can stably discover the optimal architectures of the generator and discriminator adapting to the specific dataset. The experiments in Section \ref{sec4.2}, \ref{sec4.3.2}, \ref{sec4.4} and \ref{sec4.5} support this conclusion.
		
	\end{itemize}

	Indeed, the GAN models searched by our EGANS limit the weak discrimination of their synthesized visual features.  As shown in Fig. \ref{fig:t-sne}, although the produced visual features of unseen classes are diverse, they cannot keep the promising intra-class compactness and inter-class separability. This inevitably misleads the ZSL classification, and thus limits the potential of generative ZSL models.

	\section{Conclusion}\label{sec6}
	In this paper, we proposed a novel EGANS to automatically discover the optimal GAN with desirable adaptation and stability for advancing the generative ZSL. EGANS incorporates cooperative dual evolution with respect to generator architecture search and discriminator architecture search into a unified evolutionary adversarial framework. Notably, we introduced a similar evolution algorithm for optimizing the generator/discriminator architecture search for conducting effective and stable training. The competitive results on four various datasets (\textit{e.g.}, CUB, SUN, FLO, and AWA2) demonstrate the superiority and great potential of EGANS for generative ZSL. 
	
	In the future, we will take discrimination into account during the evolution search process. For example, we can take the classification accuracy into the fitness function to evaluate the effectiveness of the evolved generator and discriminator.


	\ifCLASSOPTIONcaptionsoff
	\newpage
	\fi

	\bibliographystyle{IEEEtran}
	\bibliography{mybibfile}


\end{document}